\documentclass{article}

\usepackage[preprint]{neurips_2023}
\usepackage{microtype}
\usepackage{graphicx}
\usepackage{amsfonts}
\usepackage{subfigure}
\usepackage[dvipsnames]{xcolor}
\usepackage{amsmath}
\usepackage{graphics}
\usepackage{adjustbox}
\usepackage{bm}
\usepackage{booktabs} % for professional tables
\usepackage{makecell}
\usepackage{bm}
\usepackage{amssymb}
\usepackage[subfigure]{tocloft}
\usepackage{wrapfig}
\usepackage{floatrow}
\newfloatcommand{capbtabbox}{table}[][\FBwidth]
\usepackage[font=small,labelfont=bf,tableposition=top]{caption}
\usepackage{hyperref}
\hypersetup{
 colorlinks,
 citecolor=MidnightBlue,
 linkcolor=MidnightBlue,
 urlcolor=Blue}
\usepackage{minitoc}
\usepackage{amsthm}
\usepackage[utf8]{inputenc} % allow utf-8 input
\usepackage[english]{babel}
\usepackage[T1]{fontenc}    % use 8-bit T1 fonts
\usepackage{hyperref}       % hyperlinks
\usepackage{url}            % simple URL typesetting
\usepackage{booktabs}       % professional-quality tables
\usepackage{amsfonts}       % blackboard math symbols
\usepackage{nicefrac}       % compact symbols for 1/2, etc.
\usepackage{microtype}      % microtypography
\usepackage{xcolor}         % colors
\usepackage[mathscr]{euscript}
\usepackage{mathabx}

\newtheorem{Proposition}{Proposition}
\newtheorem{Theorem}{Theorem}
\DeclareCaptionLabelFormat{andtable}{\tablename~\thetable ~\& #1~#2}
\DeclareSymbolFont{rsfs}{U}{rsfs}{m}{n}
\DeclareSymbolFontAlphabet{\mathscrsfs}{rsfs}
\DeclareMathOperator*{\argmin}{argmin}

\newcommand{\beginsupplement}{%
  \renewcommand{\thetable}{S.\arabic{table}}%
  \renewcommand{\thefigure}{S.\arabic{figure}}%
  \renewcommand{\thesection}{S.\arabic{section}}%
  \renewcommand{\thesubsection}{S.\arabic{section}.\arabic{subsection}}%   
}

\advance\cftsecnumwidth 1em\relax
\advance\cftsubsecindent 1em\relax
\advance\cftsubsecnumwidth 1em\relax
\title{Learning Functional Transduction}

\author{%
  Mathieu Chalvidal \\
  Artificial and Natural Intelligence Toulouse Institute\\ Universite de Toulouse, France\\
  \texttt{mathieu\_chalvid@brown.edu} \\
   \And
   Thomas Serre \\
   Carney Institute for Brain Science \\
   Brown University, U.S. \\
   \texttt{thomas\_serre@brown.edu} \\
    \And
   Rufin VanRullen \\
   Centre de Recherche Cerveau \& Cognition\\  
   CNRS, Universite de Toulouse, France \\
   \texttt{rufin.vanrullen@cnrs.fr}
}

\begin{document}
\bibliographystyle{apalike}
\setcitestyle{authoryear,open={(},close={)}}

\maketitle

\begin{abstract}
Research in machine learning has polarized into two general approaches for regression tasks: Transductive methods construct estimates directly from available data but are usually problem unspecific. Inductive methods can be much more specific but generally require compute-intensive solution searches. In this work, we propose a hybrid approach and show that transductive regression principles can be meta-learned through gradient descent to form efficient \textit{in-context} neural approximators by leveraging the theory of vector-valued Reproducing Kernel Banach Spaces (RKBS). We apply this approach to function spaces defined over finite and infinite-dimensional spaces (function-valued operators) and show that once trained, the \textit{Transducer} can almost instantaneously capture an infinity of functional relationships given a few pairs of input and output examples and return new image estimates. We demonstrate the benefit of our meta-learned transductive approach to model complex physical systems influenced by varying external factors with little data at a fraction of the usual deep learning training computational cost for partial differential equations and climate modeling applications.
\end{abstract}
\setcounter{tocdepth}{-1}
% \addtocontents{toc}{\protect\setcounter{tocdepth}{-1}}

\section{Introduction}

\textbf{Transduction vs. induction} $\diamond$ In statistical learning, transductive inference \citep{vapnik2006estimation} refers to the process of reasoning directly from observed (training) cases to new (testing) cases and contrasts with inductive inference, which amounts to extracting general rules from observed training cases to produce estimates.
% \citet{vapnik2006estimation} condensed this philosophy into an imperative principle for statistical learning: \textit{"When solving a problem of interest, do not solve a more general problem as an intermediate step. Try to get the answer that you really need but not a more general one."}.Transductive machine learning \citet{vapnik2006estimation} 
The former principle powers some of the most successful regression algorithms, from $k$-Nearest Neighbors \citep{cover1967nearest} to Support Vector Machines \citep{boser1992training} or Gaussian Processes \citep{williams1995gaussian}. A major advantage of such systems is their wide applicability and straightforward construction. In contrast, deep learning research has mostly endeavored to find inductive solutions by relying on the empirical evidence that stochastic gradient descent can faithfully encode functional relationships described by large datasets into the weights of a neural network. Although generic, inductive neural learning with gradient descent is compute-intensive, necessitates large amounts of data to approximate a single functional map, and poorly generalizes outside of the training distribution  \citep{jin2020quantifying} such that a slight modification of the problem might require retraining and cause "catastrophic" forgetting of the previous solution \citep{MCCLOSKEY1989109}. This may be particularly problematic for real-world applications where data has heterogeneous sources, or only a few examples of the target function are available.\\
% Finally, they show poor robustness to adversarial attack \citep{szegedy2013intriguing} or anomalous data contamination of the training distribution \citep{zhang2021understanding}. 
% To become practically useful, such data-driven methods should also perform well outside of the data set they are trained with and should extrapolate to different parameters and contexts.\\

\begin{figure}[h]
\vspace{-1.1cm}
  \begin{center}
\includegraphics[width=\textwidth]{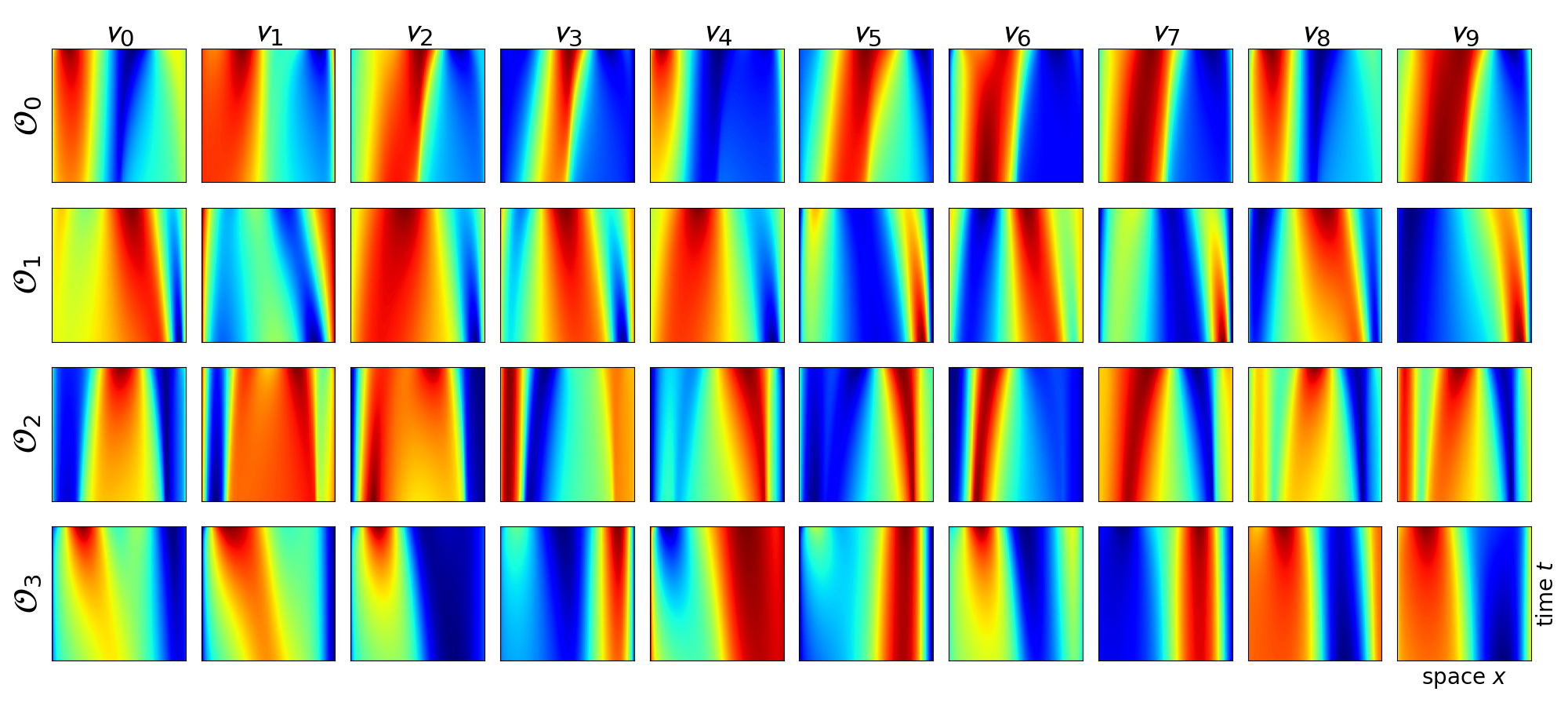}
  \end{center}
% \vspace{-0.4cm}
\vspace{-0.6cm}
\setlength{\belowcaptionskip}{-1cm}
  \caption{Batches of functional images $ \mathcal{T}_{\bm{\theta}}(\mathcal{D}_{\mathcal{O}_i})(\bm{v}_j) \approx \mathcal{O}_i(\bm{v}_{j}) = \bm{u}_{j}(x,t) \in C([0,1]^{2},\mathbb{R})$ obtained with the same \textit{Transducer} model $\mathcal{T}_{\bm{\theta}}$ but conditioned, at each row, by a different dataset $(\mathcal{D}_{\mathcal{O}_i})_{i\leq 3}$ during feedfoward computation.
  % corresponding to different functional operators $(\mathcal{O}_1,\mathcal{O}_2,\mathcal{O}_3)$, and different input functions $(\bm{v}_1,\bm{v}_2,\bm{v}_3)$.
  Each underlying operator $\mathcal{O}_i$ corresponds to a different advection-diffusion-reaction equation (defined in Sec. \ref{ADRsection}) with spatially varying advection, diffusion, and reaction parameters unseen during training, and functions $(\bm{v}_j)_{j\leq 9}$ correspond to initial conditions. \textbf{While usual neural regression approaches learn a \textit{single} target function (one row), our model learns to approximate instantaneously an \textit{infinity} of them.} }
% \vspace{-0.8cm}
\end{figure}

\textbf{Meta-learning to regress functions} $\diamond$ In this work, we meta-learn a regression program in the form of a neural network able to approximate instantaneously an infinity of functions defined on finite or infinite-dimensional spaces through a transductive formulation of the solution. Namely, our model is meta-trained to take as input any dataset $\mathcal{D}_{\mathcal{O}}$ of pairs $(\bm{v}_i,\mathcal{O}(\bm{v}_i))_{i\leq I}$ of some target function $\mathcal{O}$ together with a query element $\bm{v'}$ and produces directly an estimate of the image $\mathcal{O}(\bm{v'})$. After meta-training, our network is able to perform regression of unseen operators $\mathcal{O}'$ from varying dataset sizes in a single feedforward pass, such that our model can be interpreted as performing \textit{in-context functional learning}. In order to build such a model, we leverage the theory of Reproducing Kernel Banach Spaces (RKBS) \citep{micchelli2004function,ZHANG2013195,lin2022reproducing}
% \citep{Wang2015,kadri2016operator} 
and take inspiration from the Transformer's \citep{AttentionIsAllYouNeed} attention mechanism interpreted as a parametric vector-valued reproducing kernel. While kernel regression might be plagued by the ``curse of dimensionality'' \citep{bellman1966dynamic, aggarwal2001surprising}, we show that our meta-learning approach can escape this pitfall, allowing, for instance, to perform instantaneous regressions over spaces of operators from a few example points, by building 
solutions to regression problem instances directly from the general reproducing kernel associated with such spaces.
% From a control perspective, our method can be interpreted as building an \textit{open-loop controller} where input-output examples condition a neural operator to produce a desired functional response. 

\textbf{Contributions} $\diamond$ We introduce the \textit{Transducer}, a novel meta-learning approach leveraging reproducing kernel theory and deep learning methods to perform instantaneous regression of an infinity of functions in reproducing kernel spaces. 
\begin{itemize} 
    \item Our model learns an implicit regression program able to identify, in a single feedforward pass, elements of specific functional spaces from any corresponding collection of input-output pairs describing the target function. Such ultra-fast regression program, which bypasses the need for gradient-based training, is also general and can be applied to functions either defined on finite dimensional spaces (scalar-valued function spaces) or infinite dimensional spaces (function-valued operator spaces).
    \item In particular, we demonstrate the flexibility and efficiency of our framework for fitting function-valued operators in two PDEs and one climate modeling problem. We show that our transductive approach allows for better generalization properties of neural operator regression, better precision when relevant data is available, and can be combined with iterative regression schemes that are too expensive for previous inductive approaches, thus holding great potential to improve neural operators applicability.
    \item To the best of our knowledge, our proposal is the first to marry vector-valued RKBS theory with deep meta-learning and might also shed new light on the in-context learning abilities observed in deep attentional architectures.
\end{itemize}

% The remainder of the paper is organized as follows: in Section \ref{Related} we discuss previous related work. In Section \ref{Problem} we introduce the mathematical formulation of the functional transductive regression problem and describe more formally our model in section \ref{Transducer}. We report in section \ref{XP} experimental results in different settings. Finally, in Section \ref{Conclusion} we summarise our work and outline future research directions.

\section{Problem formulation}\label{Problem}

Let $\mathcal{V}$ and $\mathcal{U}$ be two (finite or infinite-dimensional) Banach spaces, respectively referred to as the input and output space, and let $\mathcal{B}$ a Banach space of functions from $\mathcal{V}$ to $\mathcal{U}$.
% Throughout, we assume that all considered Banach spaces $\mathcal{S}$ are uniformly convex and uniformly Fréchet-differentiable, which allows us to equip them with their compatible semi-inner product $\langle.,.\rangle_{\mathcal{S}}:\mathcal{S}\times\mathcal{S}\mapsto \mathbb{C}$, and to define their bijective dual space $\mathcal{S}^{*} := \{\bm{s}' \mapsto \langle\bm{s}',\bm{s}\rangle_{\mathcal{S}},\bm{s}\in\mathcal{S}\}$ \citep{Giles1967ClassesOS,Koehler}.
We also note $L(\mathcal{U},\mathcal{B})$ (resp. $L(\mathcal{U})$)  the set of bounded linear operators from $\mathcal{U}$ to $\mathcal{B}$ (resp. to itself). We consider the \textit{meta-learning} problem of creating a function $\mathcal{T}$ able to approximate any functional element $\mathcal{O}$ in the space $\mathcal{B}$ from any finite collection of example pairs $\smash{\mathcal{D}_{\mathcal{O}}= \{(\bm{v}_i,\bm{u}_i) \;|\; \bm{v}_i \in \mathcal{V}, \bm{u}_i=\mathcal{O}(\bm{v}_i)\}_{i\leq n}}$. 
% More precisely, denoting $\bm{\mathcal{D}}$ the set of possible realizations $\mathcal{D}_{\mathcal{O}}$ for all function $\mathcal{O}$ in $\mathcal{B}$, we are looking for functions $\mathcal{T} : \bm{\mathcal{D}} \mapsto \mathcal{B} $ such that for a given $\epsilon > 0$, we have:
% \begin{align}
%     \forall \mathcal{O} \in \mathcal{B}, \hspace{2mm} ||\mathcal{T}(\mathcal{D}_{\mathcal{O}}) - \mathcal{O}||_\mathcal{B} < \epsilon
%     % \mathcal{T}_{\theta} \;\; : \;\; & \bm{\mathcal{D}} \mapsto \mathcal{B} \mathcal{D}_{\mathcal{O}} \mapsto \mathcal{T}_{\theta}(\mathcal{D}_{\mathcal{O}}) = \hat{\mathcal{O}} \approx  \mathcal{O}
%     \label{eq:approx}
% \end{align}
% Assuming the existence of such functions $\mathcal{T}$ (in particular, this implies that all datasets $\mathcal{D}_{\mathcal{O}}$ contain enough information to approximate $\mathcal{O}$ with $\mathcal{T}$ at a given error level $\epsilon$), 
% Finding a method to build functions $\mathcal{T}$ for a variety of spaces $\mathcal{B}$ constitutes a general but complex \textit{meta-learning} problem. 
% Since $\mathcal{B}$ is a functional space, this problem requires special care: For any given dataset $\mathcal{D}_{\mathcal{O}}$, the map $\mathcal{T}$ must i) extract statistical regularities from $\mathcal{D}_{\mathcal{O}}$ to identify the function $\mathcal{O}$ and ii) provide a computable functional approximation of $\mathcal{O}$. 
A prominent approach in statistical learning is \textit{empirical risk minimization} which consists in predefining a class $\tilde{\mathcal{B}} \subset \mathcal{B}$ of computable functions from $\mathcal{V}$ to $\mathcal{U}$ and subsequently selecting a model $\tilde{\mathcal{O}}$ as a minimizer (provided its existence) of a risk function $\mathcal{L}:\mathcal{B}\times\bm{\mathcal{D}}\mapsto \mathbb{R}$:
\begin{equation}
    \mathcal{T}(\mathcal{D}_{\mathcal{O}}) \in \argmin\limits_{\tilde{\mathcal{O}}\in \tilde{\mathcal{B}}} 
    % \frac{1}{n}
    \mathcal{L}(\tilde{\mathcal{O}},\mathcal{D}_{\mathcal{O}})
    % \sum_{i\leq n} \mathcal{L}(\mathcal{O}(\bm{v}_i),\tilde{\mathcal{O}}(\bm{v}_i)) + \varphi(||\tilde{\mathcal{O}}||_{\mathcal{B}})
    \label{eq:riskmini}
\end{equation}
For instance, the procedure consisting in performing gradient-based optimization of objective \eqref{eq:riskmini} over a parametric class $\smash{\tilde{\mathcal{B}}}$ of neural networks defines implicitly such a function $\mathcal{T}$.
% where $\theta$ represents the set of hyperparameters of the optimization algorithm as well as the model settings. 
Fundamentally, this technique works by induction: It captures the statistical regularities of a single map $\mathcal{O}$ into the parameters of the neural network $\smash{\tilde{\mathcal{O}}}$ such that $\mathcal{D}_{\mathcal{O}}$ is discarded for inference.
% , provided sufficient inductive alignment of $\hat{\mathcal{O}}$ with $\mathcal{O}$, i.e. $\forall \bm{v} \in \mathcal{V}, \; \hat{\mathcal{O}}(\bm{v}) \approx \mathcal{O}(\bm{v})$.
Recent examples of gradient-based optimization of neural networks for operator regression (i.e when $\mathcal{V}$ and $\mathcal{U}$ are infinite-dimensional) are DeepOnet \citep{lu2019deeponet} or Fourier Neural Operator (FNO) \citep{li2020fourier}. As previously discussed, for every regression problem instance, evaluating $\mathcal{T}$ with these approaches requires a heavy training procedure. Instead, we show in this work that for specific spaces $\mathcal{B}$, we can meta-learn a parametric map $\mathcal{T}_{\bm{\theta}}$ that transductively approximates (in a certain functional sense) any target function $\mathcal{O} \in \mathcal{B}$ given a corresponding dataset $\mathcal{D}_{\mathcal{O}}$ such that:
% to learn a parametric map $\mathcal{T}_{\bm{\theta}}$ that directly associates to a dataset $\mathcal{D}_{\mathcal{O}}$ its corresponding target function $\mathcal{O}$ such that:
% Despite its theoretical ubiquity, such an approach is plagued with multiple problems discussed above: heavy training procedure with high sample complexity, weak robustness and adaptability to out-of-distribution samples. 
% Moreover, the function $\mathcal{L}$ is generally non-convex with respect the optimization variable, which betray the fundamental objective of equation \eqref{eq:riskmini}. 

\begin{equation}
    \forall \bm{v} \in \mathcal{V}, \;\; \mathcal{T}(\mathcal{D}_{\mathcal{O}})(\bm{v}) = \mathcal{T}_{\bm{\theta}}(\bm{v}_1,\mathcal{O}(\bm{v}_1),\dots,\bm{v}_n,\mathcal{O}(\bm{v}_n),\bm{v}) \approx \mathcal{O}(\bm{v})
    \label{eq:transduc}
\end{equation}

% through the following definition:  
% \begin{equation}
%     \mathcal{T}_{\theta}(\mathcal{D}_{\mathcal{O}}) \in \argmin\limits_{\mathcal{O}'\in \mathcal{B}'} \frac{1}{n}\sum \mathcal{L}(\bm{u}_i,\mathcal{O}'(\bm{v}_i))
% \end{equation}

\section{Vector-valued Reproducing Kernel Banach Space regression}

In order to build $\mathcal{T}_{\bm{\theta}}$,
% associating $\mathcal{D}_{\mathcal{O}}$ to its corresponding function $\mathcal{O}$, 
% and bypass the need for solving an optimization problem, 
we leverage the structure of reproducing kernel Banach spaces of functions $\mathcal{B}$ and combine it with the universal approximation abilities of deep networks. As we will see in the experimental section, RKBS are very general spaces occurring in a wide range of machine learning applications. We start by recalling some elements of the theory of vector-valued RKBS developed in \citet{ZHANG2013195}. Namely, we will consider throughout \textit{uniform} Banach spaces $\mathcal{S}$ (such condition guarantees the unicity of a compatible semi-inner product $\langle.,.\rangle_\mathcal{S}: \mathcal{S}\times \mathcal{S} \mapsto \mathbb{R}$, i.e. $\forall \bm{s} \in \mathcal{S}, \langle \bm{s},\bm{s}\rangle_\mathcal{S} = ||\bm{s}||^2_{\mathcal{S}}$ and allows to build a bijective and isometric dual space $\mathcal{S}^{*}$).
% For a more in-depth treatment of RKBS, see also \citet{lin2022reproducing}.
% \begin{equation}
%     \mathcal{O}_0^{*} = \sum_{i\leq I}(\mathcal{K}(\bm{v}_i,.)(\bm{y}_i))^{*}
% \end{equation}

% \begin{figure}[h]
%  \vskip +0.2cm
%   \begin{center}
% \includegraphics[width=0.2\textwidth]{TransducerLayer.png}
%   \end{center}
%   \vskip -5pt
%   \caption{Schema of a \textit{Transducer} layer. A collection of input and output elements $(\bm{v}^l_i,\bm{u}^l_i)$ are transformed in parallel by one-layer fully connected residual networks into intermediate representations $(\bm{\tilde{v}}^l_i,\bm{\tilde{u}}^l_i)$. Every element $(\bm{\tilde{u}}^l_j)$ is further transformed by the kernel operation $\mathcal{K}_{\theta_l}$ which extract relations between $(\bm{\tilde{v}}^l_i)$ to form the residual contribution $\sum_i\kappa_{\theta}^l(\bm{\tilde{v}}^l_i,\bm{\tilde{v}}^l_j)(\bm{\tilde{u}}^l_i)$. } 
%   \vskip -3pt
% \end{figure}

% : \mathcal{B} \times \mathcal{B}^{*} \mapsto \mathcal{L}(\mathcal{U})$ such that $\forall \; \mathcal{O},\mathcal{O}^{*} \in \mathcal{B} \times \mathcal{B}^{*}, \langle\mathcal{O},\mathcal{O}^{*}\rangle_{\mathcal{B}}=\mathcal{O}^{*}(\mathcal{O})$
% zhang2009reproducing ,JMLR:v17:11-315}.

\begin{Theorem}[\textbf{Vector-valued RKBS \citep{ZHANG2013195}}]
A $\mathcal{U}$-valued reproducing kernel Banach space $\mathcal{B}$ of functions from $\mathcal{V}$ to $\mathcal{U}$ is a Banach space such that for all $\bm{v} \in \mathcal{V}$, the point evalutation $\delta_{\bm{v}}: \mathcal{B}\mapsto \mathcal{U}$ defined as $ \delta_{\bm{v}}(\mathcal{O})=\mathcal{O}(\bm{v})$ is continuous. In this case, there exists a unique
% $\mathcal{B}^{*}$-valued kernel 
function $\mathcal{K} :\mathcal{V} \times \mathcal{V} \mapsto L(\mathcal{U})$ such that for all $(\bm{v},\bm{u}) \in \mathcal{V}\times \mathcal{U}$:
\begin{equation}
    \begin{cases}
    \bm{v}' \mapsto \mathcal{K}(\bm{v},\bm{v}')(\bm{u}) \in \mathcal{B}\\
    \forall \; \mathcal{O} \in \mathcal{B}, \; \langle\mathcal{O}(\bm{v}),\bm{u}\rangle_{\mathcal{U}} = \langle \mathcal{O},\mathcal{K}(\bm{v},.)(\bm{u})\rangle_{\mathcal{B}}\\
    % \overline{\text{span}}\{\mathcal{K}(\bm{v},.)(\bm{u})^{*},\bm{v}\in\mathcal{V},\bm{u}\in\mathcal{U}\}=\mathcal{B}^{*}\\
    \forall \; \bm{v}' \in \mathcal{V}, \; \|\mathcal{K}(\bm{v},\bm{v}')\|_{L(\mathcal{U})} \leq \|\delta_{\bm{v}}\|_{L(\mathcal{B},\mathcal{U})}\|\delta_{\bm{v}'}\|_{L(\mathcal{B},\mathcal{U})}
    \end{cases}
\end{equation}
\label{RKBS}
\end{Theorem}
\vspace{-0.2cm}
Informally, theorem (\ref{RKBS}) states that RKBS are spaces sufficiently regular such that the image of \textit{any} element $\mathcal{O}$ at a given point $\bm{v}$ can be expressed in terms of a unique function $\mathcal{K}$. The latter is hence called the \textit{reproducing kernel} of $\mathcal{B}$ and our goal is to leverage such unicity to build the map $\mathcal{T}_{\bm{\theta}}$. 
Let $\bm{\mathcal{D}}$ be the set of all datasets $\mathcal{D}_{\mathcal{O}}$ previously defined. The following original theorem gives the existence of a solution to our meta-learning problem and relates it to the reproducing kernel.

\begin{Theorem}[\textbf{RKBS representer map}]\label{representermap} 
Let $\mathcal{B}$ be a $\mathcal{U}$-valued RKBS from $\mathcal{V}$ to $\mathcal{U}$, if for any dataset $ \mathcal{D}_{\mathcal{O}} \in \bm{\mathcal{D}}$, $\mathcal{L}(.,\mathcal{D}_{\mathcal{O}})$ is lower semi-continuous, coercive and bounded below, then there exists a function $\mathcal{T}:\bm{\mathcal{D}}\mapsto\mathcal{B}$ such that 
% for any dataset $ \mathcal{D}_{\mathcal{O}} \in \bm{\mathcal{D}}$, 
$\mathcal{T}(\mathcal{D}_{\mathcal{O}})$ is a minimizer of equation \eqref{eq:riskmini}. If $\mathcal{L}$ is of the form
$\smash{\mathcal{L}(.,\mathcal{D}_{\mathcal{O}}) = \Tilde{\mathcal{L}}\circ\{\delta_{\bm{v}_i}\}_{i\leq n}}$ with $\smash{\Tilde{\mathcal{L}}:\mathcal{U}^{n}\mapsto \mathbb{R}}$, then the dual $\smash{\mathcal{T}(\mathcal{D}_{\mathcal{O}})^{*}}$ is in $\overline{span}\{\mathcal{K}(\bm{v}_i,.)(\bm{u})^{*},i\leq n,\bm{u} \in \mathcal{U}\}$. Furthermore, if for any $\mathcal{D}_{\mathcal{O}}$, $\mathcal{L}(.,\mathcal{D}_{\mathcal{O}})$ is strictly-convex, then $\mathcal{T}$ is unique.
% if $\Tilde{\mathcal{L}}$ is strictly convex, then $\mathcal{T}$ is unique.
% \begin{equation}
%     \forall \mathcal{D} \in \bm{\mathcal{D}}, \;\; 
%     \begin{cases}
%         \mathcal{T}(\mathcal{D}) = \argmin\limits_{\mathcal{O} \in \mathcal{B}}\{\|\mathcal{O}\|_{\mathcal{B}}, \text{ s.t. } \mathcal{O}(\bm{v}_i) = \bm{u}_i, i\leq n\}\\
%         \mathcal{T}(\mathcal{D})^{*} \in \overline{span}\{\mathcal{K}(\bm{v}_i,.)(\bm{u})^{*},i\leq n,\bm{u} \in \mathcal{U}\}.
%     \end{cases}
%     \label{eq:representermap}
% \end{equation}
% and we have that $\forall \mathcal{D} \in \bm{\mathcal{D}}, \;\; \mathcal{T}(\mathcal{D})^{*} =  $
\end{Theorem}

While theorem (\ref{representermap}) provides conditions for the existence of solutions to each regression problem defined by \eqref{eq:riskmini}, the usual method consisting in solving instance-specific minimization problems derived from representer theorems characterizations is generally intractable in RKBS for several reasons (non-convexity and infinite-dimensionality of the problem w.r.t to variable $\bm{u}$, non-additivity of the underlying semi-inner product). Instead, we propose to define image solutions $\smash{\mathcal{T}(\mathcal{D}_{\mathcal{O}}) = \sum_{i\leq n} \mathcal{K}_{\bm{\theta}}(\bm{v}_i,.)(\tilde{\bm{u}_i})}$ where $\mathcal{K}_{\bm{\theta}}$ and $(\tilde{\bm{u}_i})$ are respectively the learned approximation of the $\mathcal{U}$-valued reproducing kernel $\mathcal{K}$ and a set of functions in $\mathcal{U}$ resulting from a sequence of deep transformations of image examples $(\bm{u}_i)$ that we define below.

\textbf{Transformers attention as a reproducing kernel} $\diamond$ 
% In RKBS, contrary to RKHS, the adjoint can not be directly expressed from $\Phi$ itself due to the construction of the semi-inner product). 
We first need to build $\mathcal{K}$. Several pieces of work have proposed constructions of $\mathcal{K}$ in the context of a non-symmetric and nonpositive semi-definite real-valued kernel \citep{zhang2009reproducing,Georgiev,LinRongrong,xu2019generalized}. In particular, the exponential key-query function in the popular Transformer model \citep{AttentionIsAllYouNeed} has been interpreted as a real-valued reproducing kernel $\bm{\kappa}_{\bm{\theta}}:\mathcal{V} \times \mathcal{V} \mapsto \mathbb{R}$ in \citet{Wright2021TransformersAD}. We extend below this interpretation to more general vector-valued RKBS: 

\begin{Proposition}[\textbf{Dot-product attention as $\mathcal{U}$-valued reproducing kernel}]
% Let $\mathcal{U}$ and $\mathcal{V}$ be finite dimensional,
Let $(p_j)_{j\leq J}$ a finite sequence of strictly positive integers, 
% $(\mathcal{F}^{j})_{j\leq J}$ be feature spaces equipped with forms $\smash{\langle.,.\rangle_{\mathcal{F}^{j}}:\mathcal{F}^{j}\times \mathcal{F}^{j}\mapsto \mathbb{R}}$, let 
let $\smash{(A^{j}_{\bm{\theta}})_{j\leq J}}$ be applications from $\mathcal{V}\times\mathcal{V}$ to $\mathbb{R}$, let $\smash{V^{j}_{\bm{\theta}}}$ be linear applications from $L(\mathcal{U},\mathbb{R}^{p_j})$ and $W_{\bm{\theta}}$ a linear application from $\smash{L(\prod\limits_{j\leq J}\mathbb{R}^{p_j},\mathcal{U})}$, the (multi-head) application $\bm{\kappa}_{\bm{\theta}}: \mathcal{V} \times \mathcal{V} \mapsto L(\mathcal{U})$ defined by 
\begin{equation}
\bm{\kappa}_{\bm{\theta}}(\bm{v},\bm{v}')(\bm{u})  
% \Phi(\bm{v}')(\Phi^{\dag}(\bm{v}'))(\bm{u})
\triangleq W_{\bm{\theta}}\bigg(\big[...,A^{j}_{\bm{\theta}}\big(\bm{v},\bm{v}'\big)
% \exp{\big(\frac{\langle Q^{j}_{\bm{\theta}}\bm{v},K^{j}_{\bm{\theta}}\bm{v}'\rangle_{\mathbb{R}^{d_j}}}{\tau}}\big)
\cdot V^{j}_{\bm{\theta}}(\bm{u}),
...\big]_{j\leq J}\bigg)
\label{discreteattention}
\end{equation}
is the reproducing kernel of an $\mathcal{U}$-valued RKBS. In particular, if $\mathcal{U}=\mathcal{V} = \mathbb{R}^p$, for $p\in \mathbb{N}^{+}$ and $A^{j}_{\bm{\theta}}= \exp{\big(\frac{1}{\tau}(Q^{j}_{\bm{\theta}}\bm{v})^{T}(K^{j}_{\bm{\theta}}\bm{v}')}\big)/\sigma(\bm{v},\bm{v}')$ with $\smash{(Q^{j}_{\bm{\theta}},K^{j}_{\bm{\theta}})_{j\leq J}}$ applications from $L(\mathcal{V},\mathbb{R}^{d})$, $\bm{\kappa}_{\bm{\theta}}$ corresponds to the dot-product attention mechanism of \citet{AttentionIsAllYouNeed}.
\end{Proposition}
Note that in \eqref{discreteattention}, the usual softmax normalization of the dot-product attention is included in the linear operations $\smash{A^{j}_{\bm{\theta}}}$ through $\sigma$. We show in the next section how such kernel construction can be leveraged to build the map $\mathcal{T}_{\theta}$ and that several variations of the kernel construction are possible, depending on the target space $\mathcal{B}$ and applications. Contrary to usual kernel methods, our model jointly builds the full reproducing kernel approximation $\mathcal{K}_{\bm{\theta}}$ and the instance-specific parametrization $(\bm{\tilde{u}}_i)_{i\leq I}$ by integrating the solutions iteratively over several residual kernel transformations. We refer to our system as a \textit{Transducer}, both as a tribute to the Transformer computation mechanism from which it is inspired and by analogy with signal conversion devices.

% Moreover, equation \eqref{discreteattention} shows that the Transformer's attention transformer can be interpreted as the parametric approximation of a particular vector-valued RKBS.

\section{The Transducer}\label{Transducer}

\textbf{Model definition} $\diamond$ We define $\mathcal{T}_{\bm{\theta}}$ as the sum of $L$ residual kernel transformations $\{\bm{\kappa}_{\bm{\theta}}^\ell\}_{\ell\leq L}$ whose expression can be written:
\begin{equation}
\forall \; \bm{v} \in \mathcal{V},\;\; \mathcal{T}_{\bm{\theta}}(\mathcal{D}_{\mathcal{O}})(\bm{v})  = \sum\limits_{i\leq I} \mathcal{K}_{\bm{\theta}}(\bm{v}_i,\bm{v})(\tilde{\bm{u}_i})
= \sum\limits_{i\leq I}\sum\limits_{\ell\leq L}\bm{\kappa}_{\bm{\theta}}^\ell(\bm{v}^\ell_i,\bm{v}^{\ell})(\bm{u}_{i}^\ell)
\label{eq:kernel}
\end{equation}
where $(\bm{v}_i^\ell,\bm{u}_{i}^\ell)_{i\leq n,l\leq L}$ and $(\bm{v}^\ell)_{l\leq L}$ refer to sequences of representations starting respectively with $(\bm{v}_i^1,\bm{u}_i^1)_{i\leq n} = \mathcal{D}_{\mathcal{O}}$, $\bm{v}^1=\bm{v}$ and defined by the following recursive relation: 
\begin{equation}
\begin{cases}
    \bm{v}_i^{\ell+1}= F_{\bm{\theta}}^{\ell}(\bm{v}_i^\ell)\;,\;\; \bm{v}^{\ell+1}= F_{\bm{\theta}}^{\ell}(\bm{v}^\ell)\\
    % \bm{u}_i^{\ell+1} = \mathcal{T}^\ell_{\theta}\big(\{\bm{v}_i^{\ell+1},G_{\theta}^{\ell}(\bm{u}_i^\ell)\}_{i\leq I} \big)
    \bm{u}_i^{\ell+1} = \tilde{\bm{u}}_i^{\ell}+ \sum_{j}\bm{\kappa}_{\bm{\theta}}^\ell(\bm{v}_j^{\ell+1},\bm{v}_i^{\ell+1})(\tilde{\bm{u}}^\ell_j) \text{ where } \tilde{\bm{u}}_i^{\ell} = G^{\ell}_{\bm{\theta}}(\bm{u}_i^{\ell})
    % +\mathcal{T}^\ell_{\theta}\big(\bm{v}_i^{\ell+1})
\end{cases}
\label{eq:iterations}
% \Big(\bm{v}_i^{\ell+1},\bm{u}_i^{\ell+1}\Big) = \Big(\bm{v}_i^\ell + F^{\ell}(\bm{v}_i^\ell), \mathcal{T}^\ell_{\theta}\big(\bm{u}_i^\ell + G^{\ell}(\bm{u}_i^\ell)\big)\Big)
\end{equation}
where $(F_{\bm{\theta}}^\ell,G^{\ell}_{\bm{\theta}})_{\ell\leq L}$ correspond to (optional) parametric non-linear residual transformations applied in parallel to representations $(\bm{v}_i^\ell,\bm{v}_i^\ell)_{i\leq n}$ while $(\bm{\kappa}_{\bm{\theta}}^\ell)_{\ell\leq L}$ are intermediate kernel transformations of the form $\kappa: \mathcal{V} \times \mathcal{V} \mapsto \mathcal{L}(\mathcal{U})$ such as the one defined in equation \eqref{discreteattention}. Breaking down kernel estimation through this sequential construction allows for iteratively refining the reproducing kernel estimate and approximating on-the-fly the set of solutions $(\bm{\tilde{u}}_i)_{i\leq I}$. We particularly investigate the importance of depth $L$ in the experimental section.
Note that equations \eqref{eq:kernel} and \eqref{eq:iterations} allow to handle both varying dataset sizes and efficient parallel inference by building the sequences $(\bm{v}^{\ell})_{\ell \leq L}$ with $(\bm{v}^{\ell}_i)_{i\leq n,\ell \leq L}$ in batches and simply masking the unwanted cross-relational features during the kernel operations. All the operations are parallelizable and implemented on GPU-accelerated tensor manipulation libraries such that each regression with $\mathcal{T}_{\bm{\theta}}$ is orders of magnitude faster than gradient-based regression methods.  
% % noting for instance that sharing the non-linear transformations $(F^\ell,G^\ell)$ for the input and output domains can be beneficial to regression performance.
% Finally, the operator image estimate of any $\bm{v} \in \mathcal{V}$ corresponds to the sum of residual transformations: 
% \begin{equation}
% \mathcal{T}_{\theta}(\mathcal{D}_{\mathcal{O}})(\bm{v}) = \sum\limits_{\ell\leq L}\sum\limits_{i\leq I} \kappa_{\theta}^\ell(\bm{\tilde{v}}^\ell_i,\bm{v}^{\ell})(\bm{\tilde{u}}^\ell_i)
% % = \sum\limits_{i\leq I} \mathcal{K}_{\theta}(\bm{v}_i,\bm{v})(\bm{u}_i)
% \end{equation}
% where the sequence $(\bm{v}^{\ell})$  is constructed in the same way as $(\bm{v}^{\ell}_i)_{i\leq I}$ starting with $\bm{v}^{1}=\bm{v}$.

\textbf{Discretization} $\diamond$ In the case of infinite-dimensional functional input and output spaces $\mathcal{V}$ and $\mathcal{U}$, we can accommodate, for numerical computation purposes, different types of function representations previously proposed for neural operator regression and allowing for evaluation at an arbitrary point of their domain. For instance, output functions $\bm{u}$ can be defined as a linear combination of learned or hardcoded finite set of functions, as in \citet{lu2019deeponet} and \citet{Bhattacharya}. We focus instead on a different approach inspired by Fourier Neural Operators \citep{li2020fourier}, by applying our model on the $M$ first modes of a fast Fourier transform of functions $(\bm{v}_i,\bm{u}_i)_{i\leq n}$, and transform back its output, allowing us to work with discrete and finite function representations.

\textbf{Meta-training} $\diamond$ In order to train $\mathcal{T}_{\bm{\theta}}$ to approximate a solution for all problems of the form \eqref{eq:riskmini}, we jointly learn the kernel operations $(\bm{\kappa}_{\bm{\theta}}^{\ell})_{\ell \leq L}$ as well as transformations $(F_{\bm{\theta}}^\ell)_{\ell \leq L}$.
% we can construct a notion of distribution of the elements $\mathcal{O}$ in $\mathcal{B}$.(For instance, by considering a Bochner space of function over a measure space V)  
Let us assume that $\mathcal{L}$ is of the form $\smash{\mathcal{L}(\mathcal{O}',\mathcal{D}_{\mathcal{O}})=\sum_j\tilde{\mathcal{L}}(\mathcal{O}'(\bm{v}_j),\mathcal{O}(\bm{v}_j))}$, that 
% elements $\mathcal{O}$ are sampled according to a probability distribution $\beta$ over the space $\mathcal{B}$, and that for each $\mathcal{O}$,  
datasets $\mathcal{D}_{\mathcal{O}}$ are sampled according to a probability distribution $\mathfrak{D}$ over the set of possible example sets with finite cardinality and that a random variable $\mathfrak{T}$ select the indices of each test set $\mathcal{D}_{\mathcal{O}}^{\textit{test}}=\{(\bm{v}_i,\bm{u}_i) \;|\; (\bm{v}_j,\bm{u}_j) \in \mathcal{D}_{\mathcal{O}}, j \in \mathfrak{T} \}$ such that the train set is $\mathcal{D}_{\mathcal{O}}^{\textit{train}}=\mathcal{D}_{\mathcal{O}}\setminus \mathcal{D}_{\mathcal{O}}^{\textit{test}}$. Our meta-learning objective is defined as: 
\begin{equation} \mathcal{J}(\bm{\theta}) = \mathbb{E}_{\mathfrak{D},\mathfrak{T}}\Big[\sum_{j \in \mathfrak{T}} \tilde{\mathcal{L}}(\mathcal{T}_{\bm{\theta}}(\mathcal{D}_{\mathcal{O}}^{\textit{train}})(\bm{v}_j),\mathcal{O}(\bm{v}_j)) 
% + \varphi(\mathcal{T}_{\bm{\theta}}(\mathcal{D}_{\mathcal{O}}^{\textit{train}}))
\Big]
\label{eq:metaproblem}
\end{equation}
which can be tackled with gradient-based optimization w.r.t parameters $\bm{\theta}$ provided $\mathcal{L}$ is differentiable (see S.I for details). In order to estimate gradients of \eqref{eq:metaproblem}, we gather a meta-dataset of $M$ operators example sets $(\mathcal{D}_{\mathcal{O}_m})_{m\leq M}$ and form, at each training step, a Monte-Carlo estimator over a batch of $k$ datasets from this meta-dataset with random train/test splits $(\mathfrak{T}_k)$. For each dataset in the batch, in order to form outputs $\mathcal{T}_{\bm{\theta}}(\mathcal{D}_{\mathcal{O}}^{\textit{train}})(\bm{v}_j)$ defined by equation \eqref{eq:kernel}, we initialize the model sequence in \eqref{eq:iterations} by concatenating $\smash{\mathcal{D}_{\mathcal{O}}^{\textit{train}}}$ with $\smash{\mathcal{D}_{\mathcal{O}}^{\textit{query}}= \{(\bm{v}_i,0_{\mathcal{U}}) \;|\; \bm{v}_i \in \mathcal{D}_{\mathcal{O}}^{\textit{test}}\}}$ and obtain each infered output $\mathcal{T}_{\bm{\theta}}(\mathcal{D}_{\mathcal{O}}^{\textit{train}})(\bm{v}_j)$ as $\smash{\sum_{\bm{v}_i \in \mathcal{D}_{\mathcal{O}}^{\textit{train}}} \mathcal{K}_{\bm{\theta}}(\bm{v}_i,\bm{v}_j)(\tilde{\bm{u}_i})}$ . Since each regression consists in a single feedforward pass, estimating gradients of the meta-parameters $\bm{\theta}$ with respect to $\mathcal{L}$ for each batch consists in a single backward pass achieved through automatic differentiation.\\

\section{Numerical experiments}
\label{XP}

In this section, we show empirically that our meta-optimized model is able to approximate any element $\mathcal{O}$ of diverse function spaces $\mathcal{B}$ such as operators defined on scalar and vector-valued function spaces derived from parametric physical systems or regression problems in Euclidean spaces. In all experiments, we use the Adam optimizer \citep{kingma2014adam} to train for a fixed number of steps with an initial learning rate gradually halved along training. All the computation is carried on a single Nvidia Titan Xp GPU with 12GB memory. Further details can be found in S.I.
% Furthermore, to the best of our knowledge, this work is the first to show that neural networks can learn transductive regression programs on space of operators. 
% We also consider the effect of taking only P of these points for each input/output pair. For example, if we use 10\% of labeled data, we set P = bM/10c and build a training data set where each example is of the form. 

% Unless otherwise specified, we use N = 500 testing instances.

% For all experiments, we set up data in the following manner. We create N sets of input/output function pairs coming from different operators sampled in the specified operator space.

% \subsection{Online Classification in Euclidean space}

% As an introductory example 
% In this first example, we demonstrate that a deep neural network can learn a regression program in high-dimensional spaces. In this first example, we demonstrate that a deep neural network can learn to form classification decision boundary in high-dimensional spaces by meta-training our model on arbitrary classification tasks. 

% \textcolor{red}{We investigate the performance of our model when the number of labeled output function points is small. In many applications labeled output function data can be scarce or costly to obtain. Therefore, it is desirable that an operator learning model is able to be successfully trained even without a large number of output function measurements.}

\subsection{Regression of Advection-Diffusion Reaction PDEs}\label{ADRsection}

\begin{figure*}[h!]
\vskip -0.6cm
  \begin{center}
    \includegraphics[width=1.03\textwidth]{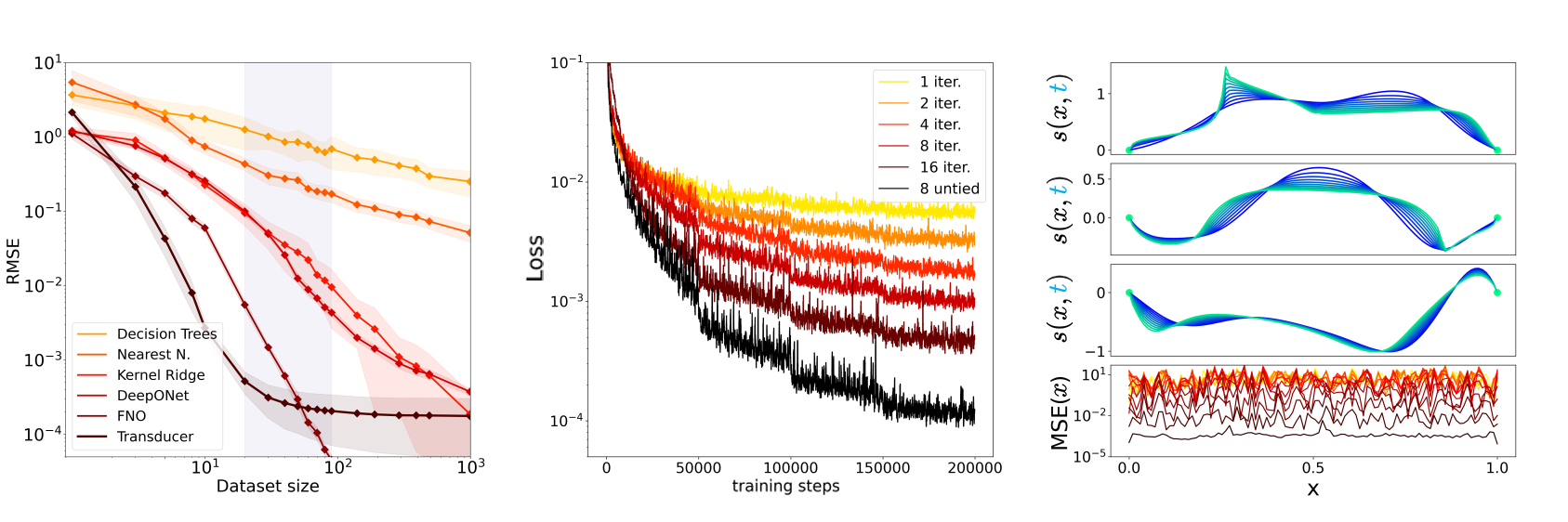}
  \end{center}
 \vspace{-0.6cm}
  \caption{\textbf{Left:} RMSEs (and 95\% C.I) on unseen operators as a function of the dataset size. The grey area corresponds to dataset cardinalities seen during the \textit{Transducer} meta-training. To provide comparison, we train baselines from scratch with the corresponding number of examples. \textbf{Middle:} Training losses of \textit{Transducers} with different depths. Applying several times the kernel improves performance, with untied weights yielding the best performance. \textbf{Right:} 3 examples of the evolution of $s(x,t)$ for different ADR equations and spatial MSEs of intermediate representations $(\bm{u}^{\ell})$ colored by iteration $\ell$. The decreasing error, consistent with the MSE reduction of deeper models, suggests that network depth allows for progressively refining function estimates.}
  \label{ADR}
%   \vskip -1pt
\end{figure*}

First, we examine the problem of regressing operators $\mathcal{O}$ associating functions $\bm{v}$  from $\mathcal{V} \subset C([0,1],\mathbb{R})$ to their solutions $\bm{u} = \mathcal{O}(\bm{v}) \subset C([0,1],\mathbb{R})$ with respect to advection-diffusion-reaction equations defined on the domain $\Omega = [0,1] \times [0,t]$ with Dirichlet boundary conditions $\bm{s}(0,t)=\bm{s}(1,t)=0$. We consider the space $\mathcal{B}$ of operators $\smash{\mathcal{O}_{(\bm{\delta}, \bm{\nu}, \bm{k},t)}}$ specifically defined by $\bm{v}(x) = \bm{s}(x,0)$, $\bm{u}(x) = \bm{s}(x,t)$ and $\bm{s}$ follows an equation
% \begin{equation}
%        \mathcal{O}_{(\bm{\delta}, \bm{\nu}, \bm{k},t)}: \bm{v} \mapsto \bm{u} \text{ such that }
%        \begin{cases}
%            \bm{v}(x) = \bm{s}(x,0) \\
%            \bm{u}(x) = \bm{s}(x,t)\\
%            \bm{s} \sim E(\bm{\delta}, \bm{\nu}, \bm{k})
%        \end{cases}
% \end{equation}
depending on unknown random continuous spatially-varying diffusion $\bm{\delta}(x)$, advection $\bm{\nu}(x)$, and a scalar reaction term $\bm{k} \sim \mathcal{U}[0,0.1]$:
\begin{equation}
\begin{split}
    \partial_t \bm{s}(x,t) =  \underbrace{\nabla\cdot(\bm{\delta}(x) \nabla_x \bm{s}(x,t))}_{\text{diffusion}} + \underbrace{\bm{\nu}(x)\nabla_x \bm{s}(x,t)}_{\text{advection}}  + \underbrace{\bm{k}\cdot(\bm{s}(x,t))^2}_{\text{reaction}}
    % + \underbrace{g(u)}_{\text{reaction}}
\end{split}
    \label{Multi}
\end{equation}
Eq. \eqref{Multi} is generic with components arising in many physical systems of interest, leading to various forms of solutions $\bm{s}(x,t)$. (We show examples for three different operators in figure \ref{ADR}.) Several methods exist for modeling such PDEs, but they require knowledge of the underlying parameters $(\bm{\delta}, \bm{\nu}, \bm{k})$ and often impose constraints on the evaluation point as well as expensive time-marching schemes to recover solutions. Here instead, we assume no \textit{a priori} knowledge of the solution and directly regress each operator $\mathcal{O}$ behavior from the example set $\mathcal{D}_{\mathcal{O}}$. 

\vspace{-0.15cm}
\renewcommand{\arraystretch}{1.3}
\begin{table}[h]
\addtolength{\tabcolsep}{+7mm}
\label{sample-table}
\begin{center}
% \begin{small}
\begin{sc}
\begin{tabular}{lrrr}
\hline
% \abovespace\belowspace
Method & RMSE & Time (s) & GFLOPs \\
\hline
% \abovespace
% Kernel Ridge  & $3\times10e^{-2}$& $\sim10^1$ & ? \\
FNO & $2.96e^{-4}$&$1.72e^{2}$ & $1.68e^{2}$\\
DeepOnet & $2.02e^{-2}$& $7.85e^{1}$ & $1.54e^{2}$\\
% \belowspace
Transducer  & $\bm{2.39e^{-4}}$ & $\bm{3.10e^{-3}}$ & $\bm{1.06e^{-1}}$         \\
 \hline
% \abovespace
% Kernel Ridge  & 95.9$\pm$ 0.2& $\sim10^1$ & ? \\
% DON & 83.3$\pm$ 0.6&$\sim 10^2$ & $\times$\\
% FNO & 83.3$\pm$ 0.6&$\sim 10^2$ & $\times$\\
% \belowspace
% Transducer    & $\bm{4.10e^{-4}}$ & $\bm{7.10e^{-1}}$ & $\bm{7.10e^{-1}}$ \\
% \hline
\end{tabular}
\end{sc}
% \end{small}
\end{center}
\vspace{-0.3cm}
\caption{RMSE and compute costs of regression over 50 unseen datasets with $n=50$ examples. Note that DeepONet and FNO are optimized from scratch while the \textit{Transducer} has been pre-trained. GFLOPs represent the total number of floating point operations for regression.}

\label{TableADR}
\end{table}
\vspace{-0.2cm}

\begin{wrapfigure}[19]{r}{0.5\textwidth}
\centering
\vskip -1cm
\includegraphics[width=\textwidth]{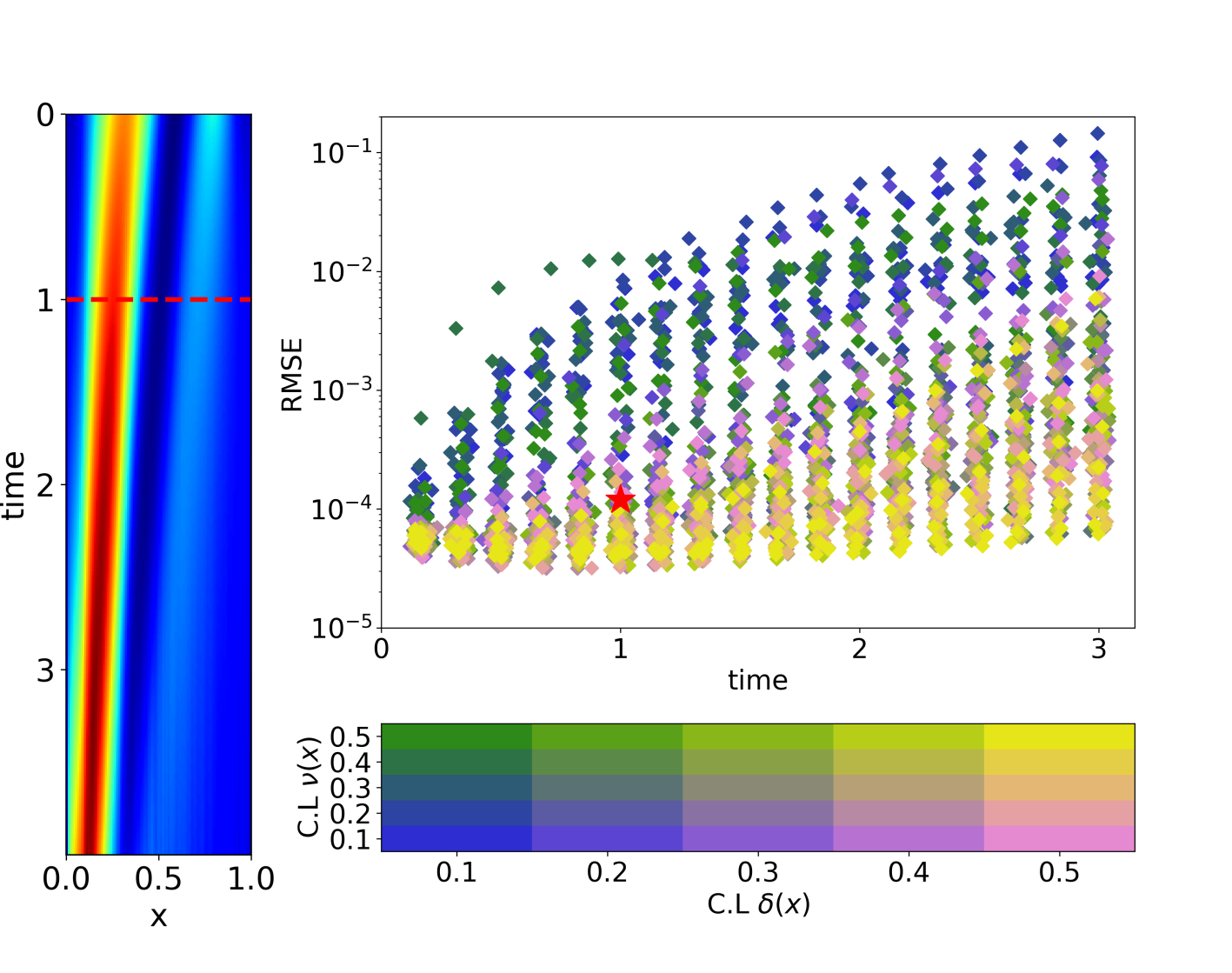}
% \vskip -0.5cm
\vspace{-0.7cm}
\caption{Example of \textit{Transducer} regression extrapolation and RMSEs on OOD tasks with $n=100$ examples. Color code corresponds to different correlation lengths used to generate the random functions $\bm{\delta}(x)$  and $\bm{\nu}(x)$. Much of the result remains below 1\% error despite never being trained on such operators.}
\label{ADRextrapolation}
\end{wrapfigure}

\textbf{Baselines and evaluation} $\diamond$ We meta-trained our model to regress $500$ different operators $\mathcal{O}_{(\bm{\delta}, \bm{\nu}, 
\bm{k},1)}$ with $t=1$ fixed and varying number of examples $n \in [20,100]$ with images evaluated at 100 equally spaced points $(x_k)_{k\in [\![0,100]\!]}$ on the domain $[0,1]$ and meta-tested on a set of $500$ operators with new parameters $\bm{\delta}, \bm{\nu}, 
\bm{k}$ and initial states $\bm{v}$. Although not directly equivalent to existing approaches, we compared our method with standard regression methods as well as inductive neural operator approximators. We applied standard finite-dimensional regression methods, $K$-Nearest-Neighbors \citep{KNN}, Decision Trees \citep{quinlan1986induction} and Ridge regression with radial basis kernel \citep{hastie2009elements} to each discretized problems $\smash{\big(\{\mathcal{O}(\bm{v}_j)(x_k)=\bm{u}_j(x_k)\}_{j,k})}$ as well as two neural-based operators to each dataset instance: DeepONet \citep{lu2021learning} and FNO \citep{li2020fourier}. For these approaches, an explicit optimization problem is solved before inference in order to fit the target operator. On the other hand, after meta-training of the \textit{Transducer}, which takes only a few minutes to converge, each regression is solved in a single feedforward pass of the network, which is orders of magnitude faster and can be readily applied to new problems (Table~\ref{TableADR}). 

\textbf{Results} $\diamond$ We first verified that our model approximates well unseen operators from the test set (Table \ref{TableADR}). We noted that our model learns a non-trivial kernel since the estimation produced with $\ell_2$-Nearest Neighbors remains poor even after $1e^3$ examples. Moreover, since our model can perform inference for varying input dataset sizes, we examined the \textit{Transducer} accuracy when varying the number of examples and found that it learns a converging regression program (Figure~\ref{ADR})
% Moreover, sheer memorization of training inputs cannot explain performance as operators lie in an infinite dimensional space with an average train distance of \textcolor{red}{XXX}.
which consistently outperforms other instance-specific regression approaches with the exception of FNO when enough data is available ($>60$). We also found that deeper \textit{Transducer} models with more layers increase kernel approximation accuracy, with untied weights yielding the best performance (figure \ref{ADR}.) 
% These results suggest that neural networks can perform online regression of such class of operator.

\textbf{Extrapolation to OOD tasks} $\diamond$ We further tested the \textit{Transducer} ability to regress different operators than those seen during meta-training. Specifically, we varied the correlation length (C.L) of the Gaussian processes used to generate functions $\bm{\delta}(x)$ and $\bm{\nu}(x)$ and specified a different target time $t'\neq1$. We showed that the kernel meta-optimized for a solution at $t=1$ transfers well to these new regression problems and that regression performance degrades gracefully as the target operators behave further away from the training set (figure \ref{ADRextrapolation}), while inductive solutions do not generalize.

\subsection{Outliers detection on 2D Burgers' equation}

\begin{figure*}[h!]
% \vskip -1pt
  \begin{center}
\includegraphics[width=\textwidth]{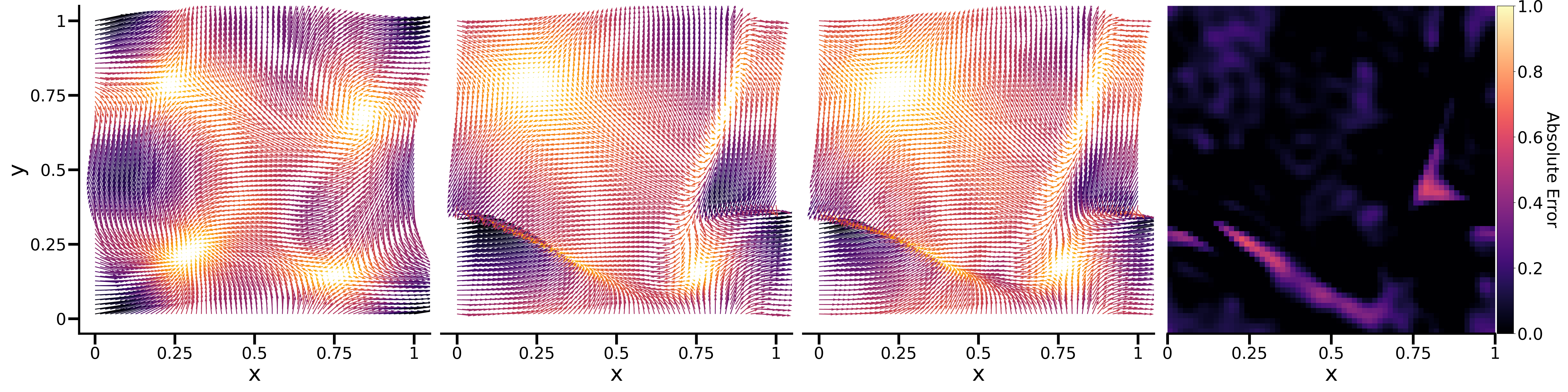}
  \end{center}
 \vskip -0.3cm
  \caption{Illustrative example of initial $(t=0)$, target $(t=10)$ and \textit{Transducer} estimation of the vector field $s(\bm{\vec{x}},t)$ discretized at resolution $64\times64$ over the domain $[0,1]^{2}$ for the Burgers' equation experiment. The last panel represents absolute error to ground truth. 
  % Different time solutions are estimated by conditioning our model with ground truth solutions at corresponding time. Despite no continuity regularization and being trained at t=1, solution estimates show remarkable smoothness property.
  }
  \label{Burger}
 \vskip -0.2cm
\end{figure*}

We further show that our regression method can fit  operators of vector-valued functions by examining the problem of predicting 2D vector fields defined as a solution of a two-dimensional Burgers' equation with periodic spatial boundary condition on the domain $\Omega = [0,1]^2\times[0,10]$:
\begin{equation}
    \partial_t \bm{s}(\bm{\vec v},t) =  \underbrace{\bm{\nu}\Delta_{\bm{v}}\cdot\bm{s}(\bm{\vec v},t)}_{\text{diffusion}} - \underbrace{\bm{s}(\bm{\vec v},t)\nabla_{\bm{x}} \bm{s}(\bm{\vec v},t)}_{\text{advection}}
    \label{eq:Burger}
\end{equation}
Here, we condition our model with operators of the form, $\bm{v}(\bm{\vec x}) = \bm{s}(\bm{\vec x},t),\bm{u}(\bm{\vec x}) = \bm{s}(\bm{\vec x},t')$
% \begin{equation}
%        \mathcal{O}_{(\bm{\nu},t,t')}: \bm{v} \mapsto \bm{u} \text{ such that }
%        \begin{cases}
%            \bm{v}(\bm{\vec x}) = \bm{s}(\bm{\vec x},t) \\
%            \bm{u}(\bm{\vec x}) = \bm{s}(\bm{\vec x},t')\\
%            \bm{s} \sim E(\bm{\nu})
%        \end{cases}
% \end{equation}
such that our model can regress the evolution of the vector field $\bm{\vec v}$ starting at any time, with arbitrary temporal increment $t'-t \leq 10$ seconds and varying diffusion coefficient $\bm{\nu} \in [0.1,0.5]$. We show in figure (\ref{Burger}) and table (\textcolor{RoyalBlue}{2}) that our model is able to fit new instances of this problem with unseen parameters $\bm{\nu}$.

\textbf{Fast and differentiable regression} $\diamond$ Since fitting with respect to training data is orders of magnitude faster than other operator regression approaches and fully differentiable, we can quickly execute expensive schemes requiring multiple regressions. This can have several applications, from bootstrapping or producing confidence intervals by varying the example set $\mathcal{D}_{\mathcal{O}}^{\textit{train}}$, or performing inverse problems using Monte-Carlo Markov Chain in the dataset space.
% We can also compute gradients with respect to training instance and perform direct optimization in the dataset space. 
We showcase an example of this potential with an outlier detection experiment: We use the \textit{Transducer} to identify outliers of a dataset of Burgers' equation with coefficient $\bm{\nu}_1$ artificially contaminated with elements from another dataset $\bm{\nu}_2>\bm{\nu}_1$ at 5$\%$ level. We identify outliers by estimating RMSEs over 5000 different regressions using random 50 $\%$ splits with outliers potentially present in both training and testing sets. This technique takes only a few seconds to estimate while outliers are clearly identified as data points with significantly higher RMSE than the dataset average (figure \textcolor{RoyalBlue}{5}). As a comparison, performing Spectral Clustering \citep{1238361} on the FFT of elements $(\bm{u}_i)$ yields very poor precision (table \textcolor{RoyalBlue}{2}) 

% \begin{figure}[h]
% \begin{center}
% \begin{floatrow}
% \ffigbox{%
%   \rule{2cm}{2cm}%
% }{%
%   \caption{A figure}%
% }
% \renewcommand{\arraystretch}{2}
% \addtolength{\tabcolsep}{0mm}
% \capbtabbox{%
% \begin{tabular}{lccc}
% \hline
% \abovespace\belowspace
%  & t=1s & t=5s & t=10s \\
% \hline
% \abovespace
% % DON & $-$&$-$ & $-$\\
% % UNET & $-$&$-$ & $-$\\
% % FNO & $8.2e^{-3}$&$1.1e^{-2}$ & $3.5e^{-2}$\\
% RMSE test  & $\bm{2.2e^{-3}}$ & $\bm{5.9e^{-3}}$ & 190\\
% \belowspace
% Outliers detec.  & $100\%$ & $100\%$ & $100\%$         \\
%  \hline
% % \abovespace
% % Kernel Ridge  & 95.9$\pm$ 0.2& $\sim10^1$ & ? \\
% % DON & 83.3$\pm$ 0.6&$\sim 10^2$ & $\times$\\
% % FNO & 83.3$\pm$ 0.6&$\sim 10^2$ & $\times$\\
% % \belowspace
% % Transducer    & $\bm{4.10e^{-4}}$ & $\bm{7.10e^{-1}}$ & $\bm{7.10e^{-1}}$ \\
% % \hline
% \end{tabular}}{
%   \caption{Relative MSE on Burgers' 2D regression set for different target times averaged over 200 different parameter conditions  $\bm{\nu} \in [0.1,0.5]$ and conditioned with 100 examples}}
% \end{floatrow}
% \end{center}
% \end{figure}

\begin{figure}[!h]
\renewcommand{\arraystretch}{1.3}
\addtolength{\tabcolsep}{0.2mm}
\begin{tabular}[b]{lccc}
\hline
% \abovespace\belowspace
% & $t$ = 1s 
 & $t$ = 5s & $t$ = 10s \\
\hline
% \abovespace
% DON & $-$&$-$ & $-$\\
% UNET & $-$&$-$ & $-$\\
% FNO & $8.2e^{-3}$&$1.1e^{-2}$ & $3.5e^{-2}$\\
RMSE (test sets)  & $\bm{2.2e^{-3}}$ & $\bm{5.9e^{-3}}$\\
% \belowspace
Outliers (Pre./Rec.) & $\bm{100\%/100\%}$  & $\bm{100\%/100\%}$\\
S.C. (Pre./Rec.)  & $6\%/85\%$ & $7\%/85\%$        \\
 \hline
 \vspace{-0.14cm}
\end{tabular}
    \centering
    \includegraphics[width=0.41\textwidth]{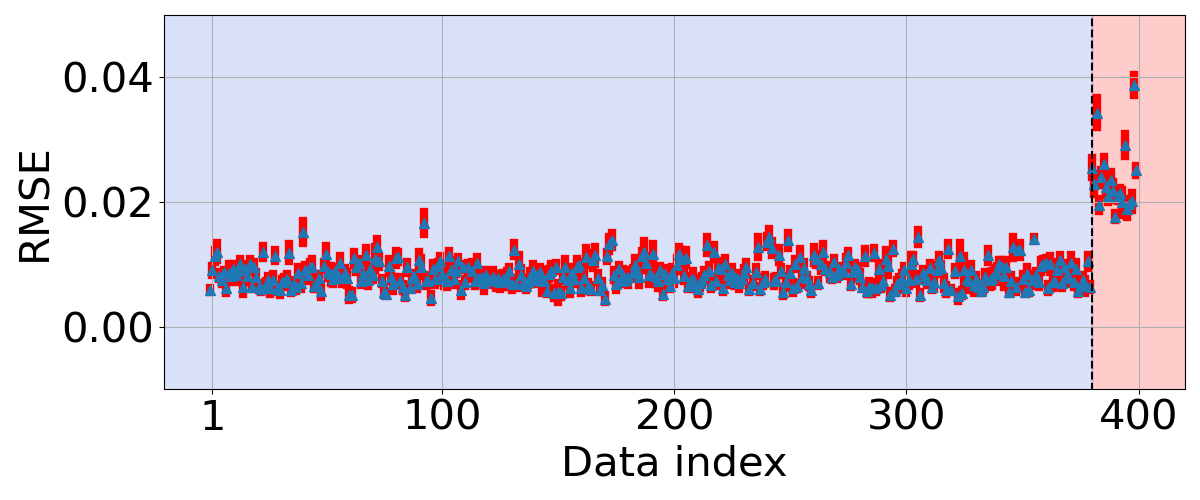}
 \vspace{-0.4cm}
    \captionlistentry[table]{A table beside a figure}
    \captionsetup{labelformat=andtable}
    \caption{\textbf{Left}: Meta-test regression and outlier detection results at two target times. RMSEs on Burgers' equations averaged over 200 different parameter conditions $\bm{\nu} \in [0.1,0.5]$ each with 100 train examples. Precision/Recall in outlier detection of the Transducer versus Spectral clustering. \textbf{Right}: RMSE distributions of each element in the contaminated dataset over the 5000 regressions. Outliers are clearly identified.}
\end{figure}

%  \begin{table}[h]
% \addtolength{\tabcolsep}{+4pt}
% \label{sample-table}
% % \vskip -0.2cm
% \begin{center}
% \begin{small}
% \begin{sc}
% \begin{tabular}{lcccr}
% \hline
% \abovespace\belowspace
% RMSE & t=1s & t=5s & t=10s \\
% \hline
% \abovespace
% % DON & $-$&$-$ & $-$\\
% % UNET & $-$&$-$ & $-$\\
% % FNO & $8.2e^{-3}$&$1.1e^{-2}$ & $3.5e^{-2}$\\
% RMSE test  & $\bm{2.2e^{-3}}$ & $\bm{5.9e^{-3}}$ & 190\\
% \belowspace
% Outliers identification  & $100\%$ & $100\%$ & $100\%$         \\
%  \hline
% % \abovespace
% % Kernel Ridge  & 95.9$\pm$ 0.2& $\sim10^1$ & ? \\
% % DON & 83.3$\pm$ 0.6&$\sim 10^2$ & $\times$\\
% % FNO & 83.3$\pm$ 0.6&$\sim 10^2$ & $\times$\\
% % \belowspace
% % Transducer    & $\bm{4.10e^{-4}}$ & $\bm{7.10e^{-1}}$ & $\bm{7.10e^{-1}}$ \\
% % \hline
% \end{tabular}
% \end{sc}
% \end{small}
% \end{center}
% % \vskip -0.2cm
% \caption{Relative MSE on Burgers' 2D regression set for different target times averaged over 200 different parameter conditions  $\bm{\nu} \in [0.1,0.5]$ and conditioned with 100 examples}
% \vskip -0.2cm
% \label{TableBurger}
% \end{table}

\subsection{Climate modeling with seasonal adaptation}

\begin{figure*}[h]
\begin{center}
\includegraphics[width=1.05\textwidth]{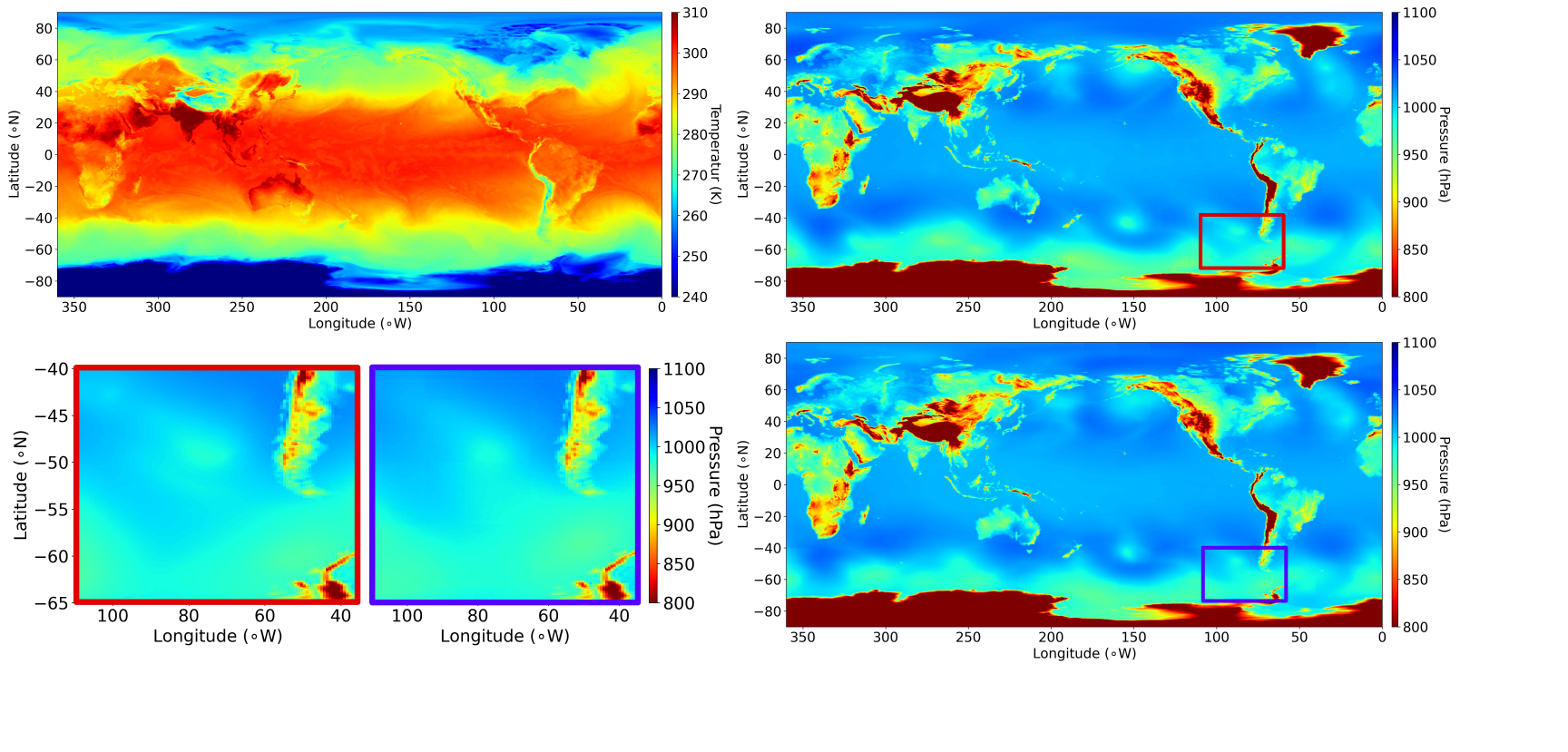}
\end{center}
\vspace{-0.9cm}
\caption{\textit{Up} - Illustrative examples of $720\times720$ temperature (\textit{left}) and pressure (\textit{right}) fields of the ERA5 dataset. \textit{Bottom} -  Estimated pressure field from conditioning the \textit{Transducer} with 15 days data dating 1 week before the target date. Insets show recovered details of the estimation (\textit{blue}) compared with ground truth (\textit{red}).}
\label{climatepanel}
% \vspace{-1cm}
\end{figure*}
One advantage of our approach is the ability to select the data that is most relevant with respect to a certain prediction task and subsequently adapt the model response. For instance, robust and precise prediction of climate variables is difficult because models need to account for seasonal variability and adapt to drifting parameters. Even with a globally large amount of available data, the underlying operator of interest might change over time or be affected by unobserved phenomena. Hence, in order to fully exploit the potential of data-driven methods, being able to capture such variations might greatly help prediction performance on fluctuating and drifting data distributions. In order to illustrate the applicability and scalability of deep transductive learning, we considered the problem of predicting the Earth's surface air pressure solely from the Earth's surface air temperature at a high resolution. Data is taken from the ERA5 reanalysis \citep{ERA5} publicly made available by the ECMWF, which consists of hourly high-resolution estimates of multiple atmospheric variables from 1979 to the current day. We model pressure estimate on a $720\times720$ grid, resulting in a spatial resolution of $0.25^{\circ} \times 0.5^{\circ}$, allowing us to capture small features such as local dynamics and geographic relief. \setlength\parfillskip{0pt}\par\setlength\parfillskip{0pt plus 1fil}
\begin{wraptable}[10]{R}{0.6\textwidth}
\renewcommand{\arraystretch}{1.2}
\addtolength{\tabcolsep}{+1pt}
\label{sample-table}
\vspace{-0.5cm}
\begin{center}
% \begin{small}
\begin{sc}
\begin{tabular}{lcccr}
\hline
% \abovespace\belowspace
Method & LWMSE (hPa) & Time (s) \\
\hline
% \abovespace
Nearest-Neighbors   & $67.326$ & $5.915$ \\
ViT   & $32.826$ & $\bm{0.053}$      \\
Transducer - (P.Y)  & $25.293$ & $0.192$ \\
% \belowspace
Transducer - (P.W)  & $\bm{22.718}$ & $0.192$ \\
\hline
\end{tabular}
\end{sc}
% \end{small}
\end{center}
\vspace{-0.3cm}
\caption{Latitude-weighted mean-square error (in hectopascals) and inference time for the earth surface pressure prediction task.}
\label{climatetable}
\end{wraptable}
\noindent Similar to \citep{Fourcastnet}, we modify a ViT backbone to incorporate a kernel transduction layer before every patch attention and compare our model to an unmodified ViT baseline with a matching number of parameters. We additionally compare with a fully transductive Nearest Neighbors approach. 
% at a latitude and longitude resolution of 0.25◦ 
In Figure \ref{climatepanel} and Table \ref{climatetable}, we present results obtained on training a Transducer with data from 2010 to 2014 and testing it on data from 2016 to 2019. We trained our model by predicting 5 random days sampled from random 20-day windows and present two test configurations: We either condition the \textit{Transducer} with a window centered at the previous year's same date (P.Y) or with a 15 days window lagging by a week (P.W) (see SI for details). Both cases outperform transductive and inductive baselines with fast inference time, confirming that our solution can scale to large problems and be combined with other deep learning modules. 

\subsection{Finite-dimensional case: MNIST-like datasets classification}

\begin{figure}[!h]
\vspace{-0.2cm}
\renewcommand{\arraystretch}{1.2}
\addtolength{\tabcolsep}{0.3cm}
\begin{sc}
\begin{tabular}[b]{lccc}
\hline
% \abovespace\belowspace
 % & $t$ = 1s 
 Method & MNIST & FashionMNIST & KMNIST \\
\hline
MAML \citep{pmlr-v70-finn17a} & 53.71\% & 48.44\% & 36.33\%\\
% LSTM  & $6\%/85\%$ & $7\%/85\%$        \\
VSML \citep{NEURIPS2021_7608de7a}  & 79.04\% &  68.49\% & 54.69\%        \\
GPICL \citep{kirsch2022generalpurpose} & 73.70 \% & 62.24\% & 53.39\%        \\
Transducer  & $\bm{81.83\%}$ & $\bm{69.85\%}$ & $\bm{60.64\%}$        \\
 \hline
\end{tabular}
\end{sc}
\vspace{-0.6cm}
\caption{Comparison of meta-test accuracies of MNIST-like datasets classification task presented in \citet{kirsch2022generalpurpose} against the Transducer.}
  \end{figure}

We finally confirm the generality of our approach in the case of finite-dimensional spaces $\mathcal{U}$ and $\mathcal{V}$ by studying the meta-learning problem presented in \citet{kirsch2022generalpurpose} which consists in regressing classification functions from the 784-dimensional space of MNIST-like images to a 10-dimensional space of one-hot class encoding (i.e functions considered are $\mathcal{O}: [0,1]^{784} \mapsto [0,1]^{10}$). We meta-train a 2-layer Transducer to classify consistently pixel-permuted and class-permuted versions of MNIST. We then meta-test the Transducer to classify the unpermuted MNIST dataset and how the regression map transfer to Fashion MNIST and KMNIST. We show that without particular fine-tuning, the Transducer outperforms previous meta-learning approaches on both the original MNIST classification task as well as Fashion MNIST and K-MNIST.

\section{Related work}\label{Related}

\textbf{Transductive Machine learning} $\diamond$ Principles of transductive statistical estimation have been formally described in \citet{gammerman1998learning,vapnik1999nature}.  Algorithms relying on relational structures between data points such as $K$-nearest neighbors \citep{cover1967nearest} and kernel methods \citep{nadaraya1964estimating,watson1964smooth} build estimates by weighing examples with respect to a certain metric space. Further, the ``kernel trick'' allows to embed possibly infinite-dimensional features \citep{ferraty2006nonparametric} 
into finite Gram matrix representations that are also well-suited for multi-task regression \citep{evgeniou2005learning,caponnetto2008universal}. Distinctively, Gaussian processes regression \citep{williams1995gaussian} combines transduction with Bayesian modeling to estimate a posterior distribution over possible functions. These techniques might suffer from the so-called ``curse of dimensionality'': with growing dimensionality, the density of exemplar point diminishes, which increases estimators' variance. More recent work combining deep learning with transductive inference has shown promising results even in high-dimensional spaces for few-shot learning \citep{snell2017prototypical,Sung_2018_CVPR} or sequence modeling \citep{jaitly2015neural}, but the vast majority of neural networks still remain purely inductive. 
% Finally, modern Hopfield networks \citep{ramsauer2020hopfield} have been shown to implement various dynamic operations on sets of representations related to the Transformer attention, offering the possibility to leverage transductive computation in the deep learning realm.

\textbf{Neural operator learning} $\diamond$
% While neural networks have primarily been applied to learning mappings between finite-dimensional spaces, their 
The universal approximation abilities of neural networks have been generalized to infinite-dimensional function spaces: \citet{chen1995universal} showed that finite neural parametrization can approximate well infinite-dimensional operators. More recent work using neural networks to perform operator regression has shown strong results \citep{lu2019deeponet}, especially when mixed with tools from functional analysis and physics \citep{raissi2017physics,li2020fourier,gupta2021multiwavelet,NEURIPS2020_4b21cf96, Nelsen2021, Wang2021,roberts2021rethinking} and constitutes a booming research direction in particular for physical applications \citep{goswami2022physics,Fourcastnet,vinuesa2022enhancing,wen2022u,pickering2022discovering}. 
% Such systems are able to approximate infinite-dimensional functional objects in a highly compressed form, and potentially offer a much more powerful substrate for learning and manipulating representations than static vectors.
Recently, the Transformer's attentional computation has been interpreted as a Petrov-Galerkin projection \citep{cao2021choose} or through Reproducing Kernel Hilbert Space theory \citep{kissas2022learning} for building such neural operators, but these perspectives apply attention to fit a single target operator.

\textbf{Meta-learning and in-context learning} $\diamond$  Promising work towards more general and adaptable machines has consisted in automatically "learning to learn" or meta-learning programs \citep{Schmidhuber1997,Vilalta}, by either explicitly treating gradient descent as an optimizable object \citep{pmlr-v70-finn17a}, modeling an optimizer as a black-box autoregressive model \citep{Ravi2017OptimizationAA} or informing sequential strategies via memorization
\citep{Santoro2016,ortega2019meta}
% Another orthogonal approach aims at meta-learning a metric over latent representations that allows for straightforward object class regression \citep{NIPS2017_cb8da676,Sung_2018_CVPR}. 
More recently, converging findings in various domains from reinforcement learning \citep{mishra2018a,laskin2022incontext}, natural language processing \citep{brown2020language,Xie2021,olsson2022context} and functional regression \citep{WhatCanTransformers}
have established the ability of set-based attentional computation in the Transformer \citep{AttentionIsAllYouNeed} for \textit{in-context} learning by flexibly extracting functional relationships and performing dynamic association such as linguistic analogy or few-shot behavioral imitation. We show that the theory of RKBS can help interpret such property and extends it to function-valued operators regression.

\section{Discussion}
\label{Conclusion}

We proposed a novel transductive model combining kernel methods and neural networks that is capable of performing regression over entire function spaces. We based our model on the theory of vector-valued Reproducing Kernel Banach Spaces and showcased several instances where it learns a regression program able, in a single feedforward pass, to reach performance levels that match or outperform previous instance-specific neural operators or meta-learning systems. Our approach holds potential to create programs flexibly specified by data and able to model entire families of complex physical systems, with particular applications in functional hypothesis testing, dataset curation or fast ensemble learning. However, one limitation is that our model relies on meta-training, which requires collecting a sufficiently diverse meta-dataset to explore the kernel space. In future work, we plan to investigate methods such as synthetic augmentation to reduce meta-training costs.

% \bibliography{biblio}

\newpage
\maketitlenoauthors
\beginsupplement
\addtocontents{toc}{\protect\setcounter{tocdepth}{2}}
\vspace{-1.2cm}
\renewcommand{\baselinestretch}{1.5}\normalsize
\tableofcontents
\renewcommand{\baselinestretch}{1.0}\normalsize
\vspace{0.2cm}
\noindent\rule{14cm}{0.4pt}

\section{Theoretical analysis}  

We propose below the proofs of the results presented in the main text. Most of the arguments are adapted from the development proposed in \citep{ZHANG2013195} which goes beyond real or complex-valued RKBS developed in \citep{Zhang2009,song2013reproducing} to develop the notion of \textit{vector-valued} RKBS. In addition, we note that assumptions regarding the properties of the RKBS of interests such as uniform Fréchet differentiability and uniform convexity have been further relaxed in other works \citep{xu2019generalized,lin2022reproducing} but are here sufficient for our discussion since they guarantee the unicity of a semi-inner product $\langle.,.\rangle_\mathcal{B}$ compatible with the norm $||.||_{\mathcal{B}}$ 
% ( i.e such that $\forall \mathcal{O} \in \mathcal{B}, \langle \mathcal{O},\mathcal{O}\rangle_\mathcal{B} = ||\mathcal{O}||_{\mathcal{B}}^2$)
\citep{Giles1967ClassesOS}. 

\subsection{Theoretical results}
\textbf{Theorem 1} $\diamond$ Theorem 1 gathers for the sake of compactness the definition of a vector-valued reproducing kernel Banach space with the properties of existence and unicity of the kernel $\mathcal{K}$. 

\begin{proof}
For any $\bm{v} \in \mathcal{V}$ and $\bm{u} \in \mathcal{U}$, the mapping $\mathcal{O}\mapsto \langle \mathcal{O}(\bm{v}),\bm{u}\rangle_\mathcal{U}$ is a bounded linear form in $\mathcal{L}(\mathcal{B})$. By Theorem 7 of  \citet{Giles1967ClassesOS}, we have the bijectivity of the duality mapping in $\mathcal{U}$, hence there exists a unique element $\mathcal{K}_{\bm{v},\bm{u}} \in \mathcal{B}$ such that: 
\begin{equation}
    \langle \mathcal{O}(\bm{v}),\bm{u}\rangle_\mathcal{U} = \langle \mathcal{O},\mathcal{K}_{\bm{v},\bm{u}}\rangle_\mathcal{B}
\end{equation}
Hence, this defines a unique function $\mathcal{K}:\mathcal{V}\times\mathcal{V}\mapsto\mathcal{L}({\mathcal{U}})$ such that:
\begin{equation}
    \forall (\bm{v},\bm{v}') \in \mathcal{V}^2,\; \forall \bm{u}\in \mathcal{U}, \;\;\;\; \mathcal{K}(\bm{v},\bm{v}')(\bm{u}) = \mathcal{K}_{\bm{v},\bm{u}}(\bm{v}')
\end{equation}
By construction $\mathcal{K}$ is unique, furthermore we have that (i) the functional $\bm{v}'\mapsto\mathcal{K}(\bm{v},\bm{v}')$ is an element of $\mathcal{B}$ (ii) it verifies the reproducing relation $ \forall (\bm{v},\bm{u}), \langle \mathcal{O}(\bm{v}),\bm{u}\rangle_\mathcal{U} = \langle \mathcal{O},\mathcal{K}(\bm{v},.)(\bm{u})\rangle_\mathcal{B}$. Finally, property (iii) follows from the following bound on the norm of $\bm{v}'\mapsto\mathcal{K}(\bm{v},.)(\bm{u})$:
\begin{equation}
    ||\mathcal{K}(\bm{v},.)(\bm{u})||_{\mathcal{B}} \leq \sup\limits_{\mathcal{O} \in \mathcal{B},||\mathcal{O}||_{\mathcal{B}\leq 1}}|\langle\mathcal{O},\mathcal{K}(\bm{v},.)(\bm{u})\rangle_{\mathcal{B}}| = \sup\limits_{\mathcal{O} \in \mathcal{B},||\mathcal{O}||_{\mathcal{B}\leq 1}}|\langle\mathcal{O}(\bm{v}),\bm{u}\rangle_{\mathcal{U}}|
    \label{eq:maj1}
\end{equation}

Further, we have by continuity of the point evaluation $\delta_{\bm{v}}:\mathcal{O} \mapsto \mathcal{O}(\bm{v})$ that: 
\begin{equation}
    |\langle\mathcal{O}(\bm{v}),\bm{u}\rangle_{\mathcal{U}}| \leq ||\delta_{\bm{v}}||_{\mathcal{L}(\mathcal{B},\mathcal{U})}.||\bm{v}||_{\mathcal{U}}
    \label{eq:maj2}
\end{equation}
Combining \eqref{eq:maj1} and \eqref{eq:maj2} allows to write:
\begin{equation}
    ||\mathcal{K}(\bm{v},\bm{v}')(\bm{u})||_{\mathcal{B}} \leq ||\delta_{\bm{v}'}||_{\mathcal{L}(\mathcal{B},\mathcal{U})}.||\mathcal{K}(\bm{v},.)(\bm{u})||_{\mathcal{B}} \leq ||\delta_{\bm{v}'}||_{\mathcal{L}(\mathcal{B},\mathcal{U})}.||\delta_{\bm{v}}||_{\mathcal{L}(\mathcal{B},\mathcal{U})}.||\bm{u}||_{\mathcal{U}}
\end{equation}
Observing that in particular for all $\bm{u}\in \mathcal{U}\textbackslash 0_{\mathcal{U}}$:
\begin{equation}
    \frac{||\mathcal{K}(\bm{v},\bm{v}')(\bm{u})||_{\mathcal{B}}}{||\bm{u}||_{\mathcal{U}}} \leq ||\delta_{\bm{v}'}||_{\mathcal{L}(\mathcal{B},\mathcal{U})}.||\delta_{\bm{v}}||_{\mathcal{L}(\mathcal{B},\mathcal{U})}
\end{equation}
concludes the proof.
% Let $\mathcal{O}$ belong to $\mathcal{B}$ and $\bm{v}$ in $\mathcal{V}$. The point evaluation $\delta_{\bm{v}}$ is in $\mathcal{L}(\mathcal{B},\mathcal{U})$ such that for all $\bm{u} \in \mathcal{U}$:
% \begin{equation}
%    |\langle \mathcal{O}(\bm{v}),\bm{u}\rangle_\mathcal{V} | \leq 
% \end{equation}
\end{proof}

\textbf{Theorem 2} $\diamond$
We first show the existence of a solution for any problem of the form \eqref{eq:riskmini} and then characterize the solution in terms of the data points. 

\begin{proof}
We first show the existence of the map $\mathcal{T}:\bm{\mathcal{D}}\mapsto \mathcal{B}$. Let us take $\mathcal{D} \in \bm{\mathcal{D}}$, by assumption the function $\smash{\mathcal{L}_{\mathcal{D}}:\Tilde{\mathcal{O}}\mapsto\mathcal{L}(\mathcal{O},\mathcal{D})}$ is weakly-lower semi-continuous, coercive and bounded below. Let us take a sequence $(\mathcal{O}_{k})_{k\in \mathbb{N}}$ of elements in $\mathcal{B}$ such that $\smash{\mathcal{L}_{\mathcal{D}}(\mathcal{O}_{k})\rightarrow \mathfrak{L}= \inf_{\mathcal{O}\in \mathcal{B}}\mathcal{L}_{\mathcal{D}}(\mathcal{O})}$. Since $\mathcal{L}_{\mathcal{D}}$ is coercive, the sequence is bounded in $\mathcal{B}$, so there is a weakly-convergent subsequence $(\mathcal{O}_{k_i})$ such that $(\mathcal{O}_{k_i}) \rightarrow \mathcal{O}_0$. Finally, by property of weakly-lower semi-continuity, we have that $\mathfrak{L}\leq \mathcal{L}_{\mathcal{D}}(\mathcal{O}_0)\leq \lim\inf \mathcal{L}_{\mathcal{D}}(\mathcal{O}_{k})$ which shows that for any $\mathcal{D}$, there exists a minimizer of $\mathcal{L}_{\mathcal{D}}$.

We now turn to the characterization of the solution $\mathcal{O}_0$ when we have that $\mathcal{L}_{\mathcal{D}}=\Tilde{\mathcal{L}}\circ \{\delta_{\bm{v}_i}\}_{i\leq n}$ with $\Tilde{\mathcal{L}}:\mathcal{U}^{n}\mapsto \mathbb{R}$. This assumption allows to exhibit a characterization of the solution in terms of annihilator and pre-annihilitors in $\mathcal{B}$ as in previous work \citep{ZHANG2013195,xu2019generalized}. Let us consider the set $S = \{\mathcal{O}\in \mathcal{B}, \mathcal{O}(\bm{v}_i) = \bm{u}_i, i\leq I\}$. It is clearly a closed convex subset of $\mathcal{B}$. Since $\mathcal{B}$ is uniformly convex, the problem
\begin{equation}
    \inf\{||\mathcal{O}||_{\mathcal{B}}, \mathcal{O}\in S \}
    \label{eq:MNI}
\end{equation} 
admits a best approximation in $S$ \citep{Megginson}. Furthermore, $\mathcal{O}_0$ is the minimizer of \eqref{eq:MNI} if and only if for all $\mathcal{O} \in S_0 = \{\mathcal{O}\in \mathcal{B}, \mathcal{O}(\bm{v}_i) = 0_{\mathcal{U}}, i\leq I\}$, we have:
\begin{equation}
    ||\mathcal{O}+\mathcal{O}_0||_{\mathcal{B}} \geq ||\mathcal{O}_0||_{\mathcal{B}} 
\end{equation}
which by \citet{Giles1967ClassesOS} is equivalent to $\mathcal{O}_0 \in (S_0)^{\perp}$. Finally, we note that $\mathcal{O}\in S_0$ if and only if
\begin{equation}
    \langle\mathcal{O},\mathcal{K}(\bm{v}_j,.)(\bm{u})\rangle_{\mathcal{B}} = \langle\mathcal{O}(\bm{v}_j),\bm{u}\rangle_{\mathcal{U}} = 0,\;\; \forall j \leq n,\; \forall \bm{u} \in \mathcal{U}
\end{equation}
which allows us to say that 
\begin{equation}
    \mathcal{O} \in \;^{\perp}\big\{(\mathcal{K}(\bm{v}_j,.)(\bm{u}))^{*}, j\leq n, \bm{u} \in \mathcal{U}\big\}
\end{equation}
Finally, we obtain the following characterization: $\mathcal{O} \in \big(^{\perp}\big\{(\mathcal{K}(\bm{v}_j,.)(\bm{u}))^{*}, j\leq n, \bm{u} \in \mathcal{U}\big\}\big)^{\perp}$. Since $\mathcal{B}$ is reflexive, we have further that $\forall S \subset \mathcal{B}, (^{\perp}S)^{\perp} = \overline{span}S$, which concludes the proof for the characterization of $\mathcal{T}(\mathcal{D})$.

Finally, if for all $\mathcal{D}$, the function $\mathcal{L}_{\mathcal{D}}$ is strictly-convex, then it guarantees the unicity of a minimizer over $\mathcal{B}$ for every problem, which in turn defines an unique map $\mathcal{T}$.    
\end{proof}

\textbf{Proposition 1} $\diamond$ The result is direct by considering the feature map characterization of vector-valued RKBS (Corollary 3.2 of \citet{ZHANG2013195}) that we recall hereafter: We first define for any linear operator $T \in \mathcal{L}(\mathcal{S}_1,\mathcal{S}_2)$ between two Banach spaces $\mathcal{S}_1,\mathcal{S}_2$, the generalized adjoint $T^{\dag} \in \mathcal{L}(\mathcal{S}_2,\mathcal{S}_1)$ as the application verifying $\langle T\bm{s},\bm{s}'\rangle_{\mathcal{S}_1}=\langle \bm{s},T^{\dag}\bm{s}'\rangle_{\mathcal{S}_2}$ for all $(\bm{s},\bm{s}')\in \mathcal{S}_1\times\mathcal{S}_2$.\\

Let $\mathcal{F}$ be a uniform Banach space and $\Phi: \mathcal{V} \mapsto \mathcal{L}(\mathcal{F},\mathcal{U})$ a feature map such that:
\begin{align}
    &\forall (\bm{v},\bm{v}') \in \mathcal{V}^2,\;\;  \Phi(\bm{v}')(\Phi^{\dag}(\bm{v})) = \mathcal{K}(\bm{v},\bm{v}') \label{eq:Kerfeat}\\
    & \overline{span}\{(\Phi^{\dag}(\bm{v})(\bm{u}))^{*},\bm{v} \in \mathcal{V},\bm{u} \in \mathcal{U}\}=\mathcal{F}^{*}\label{eq:span}
\end{align}
with $\Phi^{\dag}:\mathcal{V} \mapsto \mathcal{L}(\mathcal{U},\mathcal{F})$ is defined by: $ \forall \bm{v},\; \Phi^{\dag}(\bm{v}) = (\Phi(\bm{v}))^{\dag}.$
then the vector space $\Tilde{\mathcal{B}}=\{\Phi(.)(\bm{w})|\bm{w}\in \mathcal{F}\}$ endowed with the norm $||\Phi(.)(\bm{w})||_{\Tilde{\mathcal{B}}}$ compatible with the following semi-inner product:
\begin{equation}
\langle\Phi(.)(\bm{w}),\Phi(.)(\bm{w}')\rangle_{\mathcal{B}}=\langle\bm{w},\bm{w}'\rangle_{\mathcal{F}}
\end{equation}
is a $\mathcal{U}$-valued RKBS with reproducing kernel $\mathcal{K}$ given in \eqref{eq:Kerfeat}.
\begin{proof} 
We show our result in the case J=1 and can be directly extended to any cardinality J. By hypothesis, $\mathcal{V}$ and $\mathcal{U}$ are a uniform Banach space and so is $\mathcal{L}(\mathcal{V},\mathcal{U})$. We hence define the feature map $\Phi$ as defined by equations \eqref{eq:Kerfeat} and \eqref{eq:span} $\Phi$ and  noting here that $\mathcal{F}=\mathcal{L}(\mathcal{V},\mathcal{U})$:
\begin{align}
    \Phi :\; & \mathcal{V} \mapsto \mathcal{L}(\mathcal{L}(\mathcal{V},\mathcal{U}),\mathcal{U})\\
    & \bm{v} \mapsto \Phi(\bm{v}) = \bigg(\bm{l} \mapsto 
    \bm{l}(\bm{v})\bigg)
    % W_{\bm{\theta}}\big(A^{1}_{\bm{\theta}}\big(\bm{v},\bm{v}'\big)
% \exp{\Big(\frac{\langle Q_{\bm{\theta}}(\bm{v}),K_{\bm{\theta}}(\bm{v}')\rangle_{\mathbb{R}^{d_j}}}{\tau}}\Big)
% \odot V_{\bm{\theta}}(\bm{u})\big)\bigg)
\end{align}
In particular, by considering the uniform space $\Tilde{\mathcal{F}} = \{\bm{l} \in \mathcal{L}(\mathcal{V},\mathcal{U}) \;|\; \exists \; \bm{v}' \in \mathcal{V},\bm{u} \in \mathcal{U} \;\; \bm{l} = W_{\bm{\theta}}\big(A^{1}_{\bm{\theta}}\big(.,\bm{v}'\big).V_{\bm{\theta}}(\bm{u})\big)\} \subset \mathcal{F}$, we have the following relation:
\begin{equation}
    \forall (\bm{v},\bm{l}) \in \mathcal{V}\times\Tilde{\mathcal{F}}, \;\; \exists \; \bm{v}' \in \mathcal{V},\bm{u} \in \mathcal{U}\;\; \text{s.t} \;\;  \Phi(\bm{v})(\bm{l}) = W_{\bm{\theta}}\big(A^{1}_{\bm{\theta}}\big(\bm{v},\bm{v}'\big).V_{\bm{\theta}}(\bm{u})\big),
    % \langle \Phi(\bm{v})(\bm{l}), \bm{u}\rangle_{\mathcal{U}} = \langle W_{\bm{\theta}}\big(A^{1}_{\bm{\theta}}\big(\bm{v},\bm{v}'\big).V_{\bm{\theta}}(\bm{u}')\big),\bm{u}\rangle_{\mathcal{U}} = \langle \bm{l}, \Phi^{\dag}(\bm{v})(\bm{u})\rangle_{\mathcal{V}}
\end{equation}
identifying the adjoint $\Phi^{\dag}(\bm{v}): \mathcal{U} \mapsto \mathcal{L}(\mathcal{V},\mathcal{U})$ as  $\Phi^{\dag}(\bm{v}) : \bm{u} \mapsto \big( \bm{v'} \mapsto W_{\bm{\theta}}\big(A^{1}_{\bm{\theta}}\big(\bm{v},\bm{v}'\big).V_{\bm{\theta}}(\bm{u})\big)\big)$ and verifying the kernel relation: 
\begin{equation}
    \forall (\bm{v},\bm{v}') \in \mathcal{V}^2,\;\;  \Phi(\bm{v}')(\Phi^{\dag}(\bm{v})) = \mathcal{K}(\bm{v},\bm{v}') = W_{\bm{\theta}}\big(A^{1}_{\bm{\theta}}\big(\bm{v},\bm{v}'\big).V_{\bm{\theta}}(.)\big)
\end{equation}
Furthermore, by bijectivity of the duality map on $\Tilde{\mathcal{F}} \subset \mathcal{L}(\mathcal{V},\mathcal{U})$ that $\overline{span}\{(\Phi^{\dag}(\bm{v})(\bm{u}))^{*},\bm{v} \in \mathcal{V},\bm{u} \in \mathcal{U}\}=\Tilde{\mathcal{F}}^{*}$. The application of the feature map characterization of $\mathcal{K}$ on $\Tilde{\mathcal{F}}$ allows to conclude.

\end{proof}

\newpage

\section{Numerical implementation}

\subsection{Loss functions and evaluations}

\textbf{Definition of loss function} $\diamond$ In the case of operator regression, we meta-train models with respect to the Mean-Squarred error (MSE) over $I$ test pairs  $(\bm{v}_i,\bm{u}_i)_{i\leq I}$ of the meta-train set and $K$ evaluation points $(x_k)_{k\leq K}$ of the domain of the output functions in $\mathcal{V}$:

\begin{equation} \mathcal{L}(\tilde{\mathcal{O}},\mathcal{D}_{\mathcal{O}})  = \frac{1}{I}\sum_{i\leq I}\tilde{\mathcal{L}}(\tilde{\mathcal{O}}(\bm{v}_i),\bm{u}_i)= \frac{1}{I.K}\sum_{i\in}\sum_{k\leq K}||\tilde{\mathcal{O}}(\bm{v}_i)(x_k)-\bm{u}_i(x_k)||^2_2
    \label{eq:MSEdef}
\end{equation}
In the case of experiment 1 (ADR equation), $(x_k)_{k\leq K}$ corresponds to equally spaced points $(x_k)_{k\in [\![0,100]\!]}$ on the domain $[0,1]$. For experiment 2 (2D Burgers equation), $(x_k)_{k\leq K}$ corresponds to uniform 2D mesh $(x_{k,p})_{k\in [\![0,64]\!],p\in [\![0,64]\!]}$ discretizing the domain $[0,1]\times[0,1]$. For experiment 3 (Climate modeling), as stated in the main text, $(x_k)_{k\leq K}$ corresponds to 2D mesh $(x_{k,p})_{k\in [\![0,720]\!,p\in [\![0,720]\!]]}$ spanning the domain $[0,180^{\circ}]\times[0,360^{\circ}]$. Finally for the final finite-dimensional experiment (MNIST-like datasets), evaluation points $(k)_{k\in [\![0,10]\!]}$ corresponds to indices of 10-dimensional vectors of one-hot class encodings such that $\mathcal{L}$ corresponds to:

\begin{equation} \mathcal{L}(\tilde{\mathcal{O}},\mathcal{D}_{\mathcal{O}}) = \frac{1}{I.K}\sum_{i\in}\sum_{k\leq 10}|\tilde{\mathcal{O}}(\bm{v}_i)(k)-\bm{u}_i(k)|^2
    \label{eq:MSEMNIST}
\end{equation}

\textbf{Definition of RMSE} $\diamond$ Similarly, in the case of operator regression, we report average Relative Mean-Squarred Errors (RMSEs) defined as:

\begin{equation} \text{RMSE}(\tilde{\mathcal{O}},\mathcal{D}_{\mathcal{O}}) = \frac{1}{I.K}\sum_{i\in}\sum_{k\leq K}\frac{||\tilde{\mathcal{O}}(\bm{v}_i)(x_k)-\bm{u}_i(x_k)||^2_2}{||\bm{u}_i(x_k)||^2_2}
    \label{eq:RMSEdef}
\end{equation}

Note that for meta-training and meta-evaluation, MSEs and RMSEs are further averaged over batches of $J'$ elements $(\mathcal{O}_j)_{j\in J'}$.

\subsection{Discussion on multi-head reproducing kernels}
\textbf{Kernel definition} $\diamond$ In coherence with \citet{Wright2021TransformersAD}, we show that different expressions of the kernel $\bm{\kappa}_{\bm{\theta}}$ can be proposed. Specifically, we tested three expressions:
\begin{itemize}
    \item Exp. dot product: $A_{\bm{\theta}}(\bm{v},\bm{v}') = \exp(\frac{K_{\bm{\theta}}(\bm{v}))^T(Q_{\bm{\theta}}(\bm{v}'))}{\tau})$
    \item RBF: $A_{\bm{\theta}}(\bm{v},\bm{v}') = \exp(\frac{||K_{\bm{\theta}}(\bm{v})-Q_{\bm{\theta}}(\bm{v}')||^2_2}{\tau})$
    \item $\ell_2$-norm: $A_{\bm{\theta}}(\bm{v},\bm{v}') = ||K_{\bm{\theta}}(\bm{v})-Q_{\bm{\theta}}(\bm{v}')||^2_2$
\end{itemize}
Note that for each kernel expression, we still perform a normalization operation $\bm{v} \mapsto \frac{A_{\bm{\theta}}(\bm{v},\bm{v}_i)}{\sum_{i\leq I}A_{\bm{\theta}}(\bm{v},\bm{v}_i)}$ over the entire set $(\bm{v}_i)_{i\leq I}$ without loss of generality. We report below regression RMSE for the ADR experiment with the different expressions for the linear function $A_{\bm{\theta}}(\bm{v},\bm{v}')$ for different dataset sizes. The two first expressions yield similar result in the ADR experiment at an equal compute cost. For coherence, we present all other results with the "exponentiated dot product" kernel definition.

\begin{table}[h]
\addtolength{\tabcolsep}{+3.5pt}
\label{sample-table}
% \vskip -0.3cm
\begin{center}
\begin{small}
\begin{sc}
\begin{tabular}{|c|c|c|c|}
\hline
Kernel expression & s=10 & s=100 & s=500 \\
\hline
% \abovespace
Exp. dot product   & $2.71e-3$ & $2.39e-4$ & $1.79e-4$ \\
\hline
RBF & $8.71e-3$	& $3.46e-4$	& $3.22e-4$\\ 
\hline
$\ell_2$-norm &  $1.71e-2$ &  $6.98e-4$ & $7.33e5$ \\
\hline
\end{tabular}
\end{sc}
\end{small}
\end{center}
% \vskip -0.5cm
\caption{Results from variation of the \textit{Transducer} kernel constructions in the ADR experiment. Note that contrary to other definitions, the $\ell_2$-based kernel does not generalize to dataset cardinalities beyond those seen in the meta-training set.}
\label{hypertable}
\end{table}

\subsection{Details on model hyperparameters and architecture}

\textbf{Discretization} $\diamond$ As mentionned in the main text, in order to manipulate functional data, our model can accomodate previous forms of discretization. We particularly tested two different forms of discretization discussed in \citep{li2020fourier} and \citep{lu2019deeponet}.
\begin{itemize}
    \item In most of our experiments, we apply the \textit{Transducer} model after performing a Fast Fourier transforms (FFT) of the considered input and output functions, and transform the \textit{Transducer}'s output back to form estimates at arbitrary resolution. More specifically, we apply our model on the $d$-dimensional finite vector formed by the first modes of the Fourier transform, and discard the rest of the function spectrum. For experiments with 2D fields, we describe more precisely in section \ref{BurgerSI} how we combine the 2D FFT with our model.
    \item We also tried a 'branch' and 'trunk' networks formulation of the model as in DeepONet \citep{lu2019deeponet}. Specifically, the branch network $g:\mathcal{V}\mapsto K^P$ correspond to the \textit{Transducer} network which outputs the weight parameters $(k_p)_{p\leq P}$ for the functional basis learned by the 'trunk' networks  $f:D\mapsto K^P$ where $D$ corresponds to the domain of $\mathcal{U}$. Hence, the transducer model reads:
    \begin{equation}
        \forall \bm{x} \in D\;\; \mathcal{T}(\mathcal{D}_{\mathcal{O}})(\bm{v})(\bm{x}) = \sum\limits_{p\leq P}g_p(\mathcal{D}_{\mathcal{O}})(\bm{v}).f_p(\bm{x})
    \end{equation}
    We tested this approach in the ADR experiment by directly feeding the functions values $(\bm{v}_i(x_k))_{k\leq 100}$ and $(\bm{u}_i(x_k))_{k\leq 100}$ of the uniformly discretized domain of $\mathcal{V}$ and $\mathcal{U}$. We noted that performance was slightly worse than the Fourier method as we did not perform additional tuning such as feature augmentation for the branch network. For coherence, we kept the Fourier transform for the other experiments. 
\end{itemize}

\textbf{Feedfoward networks definition} $\diamond$ For $F_{\theta}^{\ell}$ and $G_{\theta}^{\ell}$, we use a simple feedfoward network architecture defined as Layer normalization \citep{ba2016layer} followed by one layer perceptron with GeLU activation and did not performed architectural search on this part of the network. \\

\textbf{Architecture hyperparamters} $\diamond$ We present in the following table the particular architectural choices for each experiment.

\begin{table}[h]
\addtolength{\tabcolsep}{+3.5pt}
\label{sample-table}
% \vskip -0.3cm
\begin{center}
\begin{small}
\begin{sc}
\begin{tabular}{|c|c|c|c|c|c|}
\hline
Experiment & Depth & MLP dim & dim $d$ & $\#$heads & dim heads  \\
\hline
% \abovespace
ADR   & 1-16 & 100 & 50 & 32 & 16 \\
\hline
Burgers & 10  & 800 & 800 & 64 & 16\\ 
\hline
Climate &  6  & 512 & 512 & 40 & 16\\ 
\hline
MNIST &  2  & 256 & 784 & 32 & 32\\ 
\hline
\end{tabular}
\end{sc}
\end{small}
\end{center}
% \vskip -0.5cm
\caption{Summary of the architectural hyperparameters used to build the \textit{Transducer} in the four experiments. 'Depth' corresponds to network number of layers, 'MLP dim' to the dimensionality of the hidden layer representation in $F_{\theta}^{\ell}$ and $G_{\theta}^{\ell}$, $d$ to the dimension of the discrete function representations. }
\label{hypertable}
\end{table}

\subsection{Details on meta-training}

As stated, we used for all experiments, the same meta-training procedure. We optimized \textit{Transducer} models using the Adam optimizer \citep{kingma2014adam} for a fixed number of epochs with learning rates halved multiple times across meta-training. 

\begin{table}[h]
\addtolength{\tabcolsep}{+3.5pt}
\label{sample-table}
% \vskip -0.3cm
\begin{center}
\begin{small}
\begin{sc}
\begin{tabular}{|c|c|c|c|}
\hline
Experiment & $\#$ of Epochs & learning rate &  dim heads  \\
\hline
% \abovespace
ADR   & 200 & $1e-4$ & 50  \\
\hline
Burgers & 200  & $1e-4$ & 800 \\ 
\hline
Climate &  200  & $1e-4$ & 512\\ 
\hline
MNIST &  500  & $1e-4$ & 784 \\ 
\hline
\end{tabular}
\end{sc}
\end{small}
\end{center}
% \vskip -0.5cm
\caption{Summary of the meta-learning hyperparameters used to meta-train the \textit{Transducer} in our four experiments.  }
\label{learningtable}
\end{table}

\newpage
\section{Experiments}

In this section, we provide additional details with respect to data generation and model evaluation for each experiments discussed in section (\textcolor{RoyalBlue}{5}) of the main text. 

\subsection{Advection-Diffusion-Reaction operators}
\textbf{Data generation} -- For our experiment, we collect a meta-dataset of $N=500$ datasets of the advection-diffusion-reaction trajectories on the domain $\Omega = [0,1]\times[0,1]$ by integrating the following equations: 
\begin{equation}
\forall n \in [\![1,500]\!], \quad
    \partial_t \bm{s}(x,t) = \underbrace{\nabla\cdot(\bm{\delta}_n(x) \nabla_x \bm{s}(x,t))}_{\text{diffusion}} + \underbrace{\bm{\nu}_n(x)\nabla_x \bm{s}(x,t)}_{\text{advection}} + \underbrace{\bm{k}_n\cdot(\bm{s}(x,t))^2}_{\text{reaction}}
    \label{Multiannex}
\end{equation}
We use an explicit forward Euler method with step-size $1e^{-2}$, storing all intermediate solutions on a spatial mesh of $100$ equally spaced points. Hence, our discretized reference trajectories are of dimensions $100\times100$. For each operator $\mathcal{O}_n$ we generate spatially varying diffusion and advection coefficients as random function $\bm{\delta}_n(x):[0,1] \mapsto \mathbb{R}$ and $\bm{\nu}_n(x):[0,1] \mapsto \mathbb{R}$ as well as a random scalar reaction coefficient $\bm{k}_n$. Defining $\mathcal{G}(0,k_l(x_1,x_2))$ the one-dimensional zero-mean Gaussian random field with the covariance kernel:
\begin{equation}
   k_l(x_1,x_2) = e^{\frac{-\|x_1-x_2\|^2}{2l^2}} 
\end{equation} 
and lenght-scale parameter $l=0.2$, as well as a boundary mask function $m:[0,1] \mapsto [0,1], m(x) = 1 - (2x-1)^{10}$ (to comply with Dirichlet boundary condition and preserve numerical computation stability),  we sample $\bm{\delta}_n(x)$ and $\bm{\nu}_n(x)$ according to the following equations:
\begin{itemize}
    \item \textbf{diffusion } $\bm{\delta}_n(x) = 0.01 \times u_n(x)^{2} \times m(x)$ where $u_n\sim\mathcal{G}(0,k_{0.2}(x1,x2))$
    \item \textbf{advection} $\bm{\nu}_n(x) = 0.05 \times y_n(x) \times m(x)$ where $y_n\sim\mathcal{G}(0,k_{0.2}(x1,x2))$
    \item \textbf{reaction} $\bm{k}_n \sim \mathcal{U}([0,0.3])$.
\end{itemize}

\begin{figure}[h!]
\centering
\includegraphics[width=\textwidth]{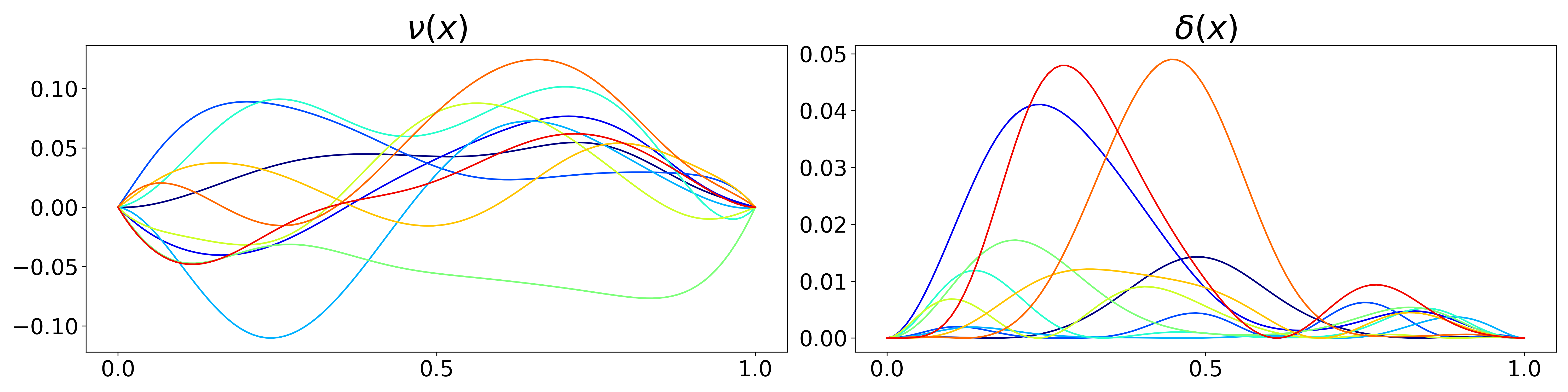}
\caption{Examples of sampled functions $\bm{\delta}(x)$ and $\bm{\nu}(x)$ used to build operators $\mathcal{O}_n$.}
% \vskip -0.2cm
\label{ADRSI}
\end{figure}

Furthermore, we collect for each dataset $i=100$ trajectories with each different initial state $s(x,0) = \bm{v}_i(x)$, where functions $\bm{v}_i(x)$ are sampled according to the following:
\begin{itemize}
    \item \textbf{initial state} $\bm{v}_i(x)$ =  $m(x) \times u_i(x)$ where $u_i\sim\mathcal{G}(0,k_{0.2}(x1,x2))$.
\end{itemize}
For meta-testing, we sample $N=500$ new datasets of the same generic advection-diffusion-reaction equation with new parameters $\bm{\delta}_n(x), \bm{\nu}_n(x), \bm{k}_n(x)$, for up to 1000 different initial states $\bm{v}_i(x)$. We present below example of function profiles present in the meta-datasets.

\begin{figure}[h!]
\centering
\includegraphics[width=0.9\textwidth]{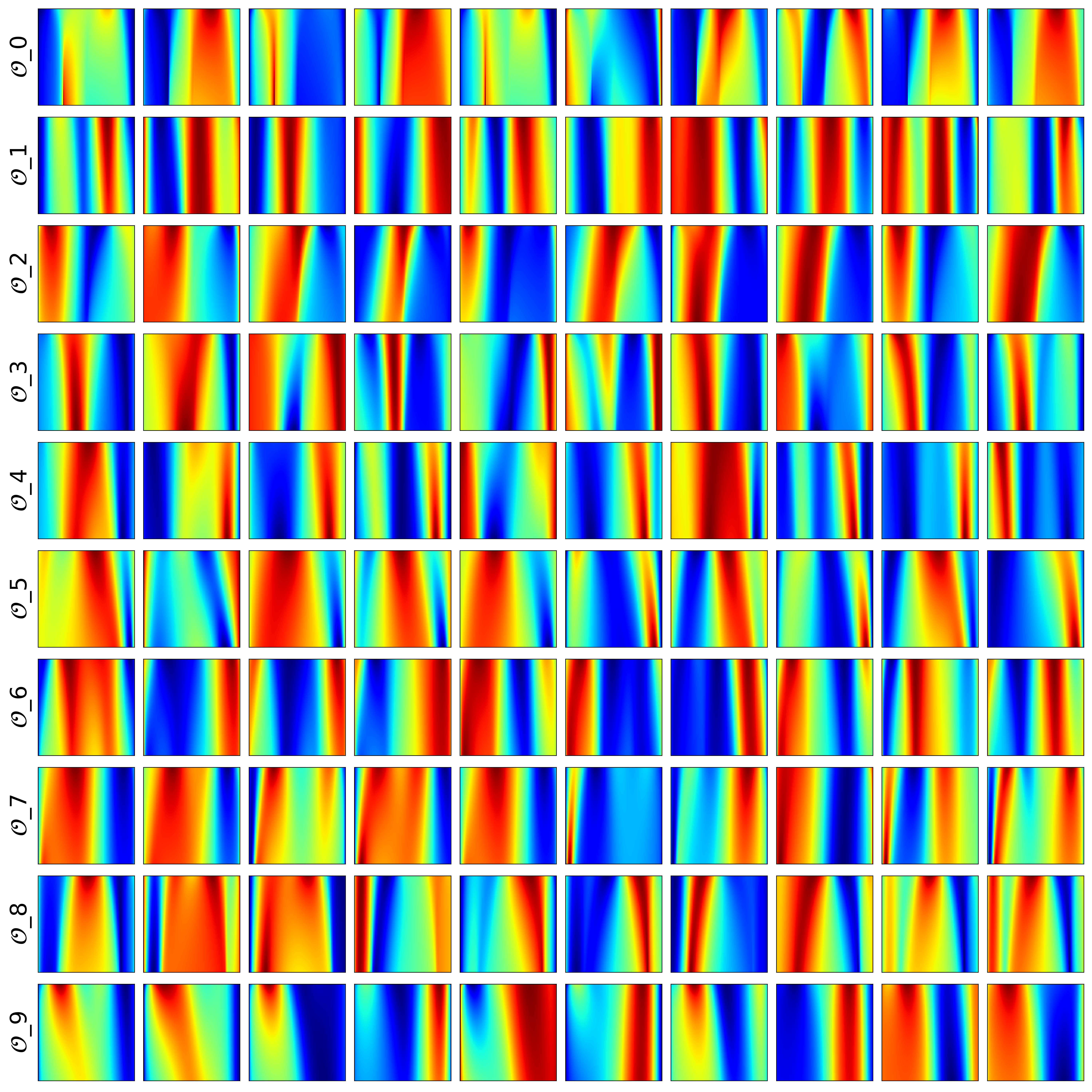}
\caption{Examples of advection-diffusion-reaction datasets (different operators by row) present in the meta-test set.}
% \vskip -0.2cm
\label{ADRSI}
\end{figure}

\textbf{Training} $\diamond$ We train Tranducers for 200K gradient steps. At each training step, we randomly draw a single operator $\mathcal{O}_n$ from the  meta-training set and isolate the pairs $(\bm{v}_i,\bm{u}_i)_{i \leq I} = (s_i(x,0),s_i(x,1))_{i \leq I}$ to form the set $\mathcal{E}_{\mathcal{O}_n}$. We  sample a "query" subset $\mathcal{Q}$ of $J=10$ pairs from $\mathcal{E}_{\mathcal{O}_n}$ to be regressed and form the input to our model by concatenating pairs of the query set $\mathcal{Q}$ (with output elements $(\bm{u}_i)_{i\in \mathcal{Q}}$ set to zero), with a non-overlaping set of $I \in [\![20,100]\!]$ example elements drawn from $(\bm{v}_i,\bm{u}_i)_{i \notin \mathcal{Q}}$. We train our model to minimize the sum of $L_2$ error between each output function of the set $Q$ and its corresponding ground truth $\bm{u}(x)=\mathcal{O}_n(\bm{v})(x)=\bm{s}(x,1)$ at the $100$ discretized positions. 

\textbf{Baselines} -- In order to implement the baseline regression algorithms, we use the scikit-learn library \citep{scikit-learn} for decisions trees, $K$-nearest neighbours and Ridge regression. We specifically tuned Ridge regression using cross-validation and selected the best-performing \textsc{'rbf'} kernel with regularisation $lambda=1e^{-3}$. For FNO \citep{li2020fourier}, we use the official PyTorch implementation provided by authors and defined for each regression, a 4-layer deep 1-dimensional FNO network with 16 modes and 64-dimensional $1\times1$ convolutions. For DeepOnet \citep{lu2019deeponet}, we implement our own PyTorch version with 4 hidden layers of 50 hidden units with ReLU activation for the branch and trunk networks.\\

\textbf{Extrapolation experiment} -- In this task, we modify the generative process of the considered operators by changing the lenght-scale parameter $l$ used to produce functions $\bm{\delta}(x)$ and $\bm{\nu}(x)$, as well as the target time $t$ used to define the operator output.
\begin{figure}[h!]
\centering
\includegraphics[width=0.9\textwidth]{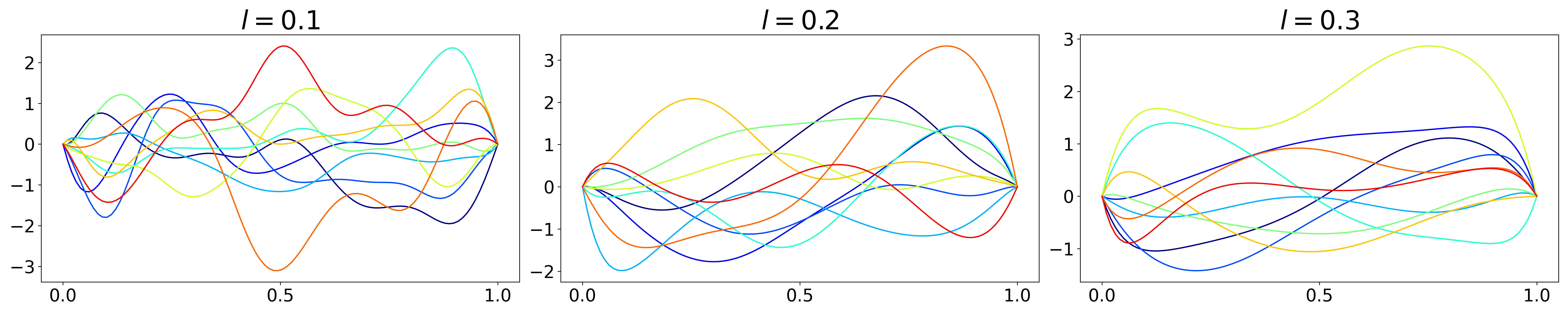}
\caption{Examples of the spatial function sampled with carying lenght scale parameter $l\in [0.1,0.2,0.3]$}
% \vskip -0.2cm
\label{ADRSI2}
\end{figure}

\begin{figure}[h!]
\centering
\includegraphics[width=0.9\textwidth]{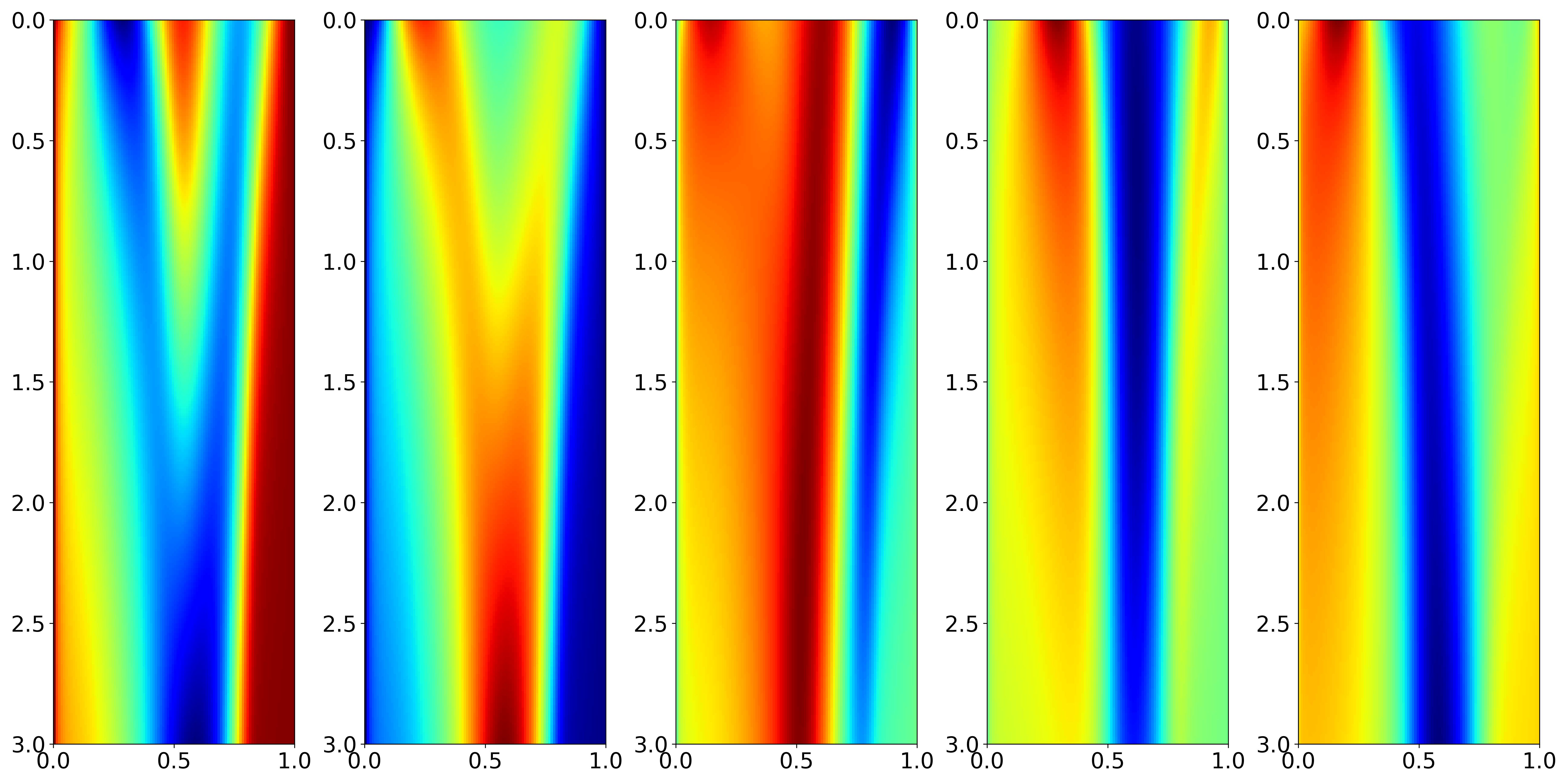}
\caption{Examples of ADR state evolution forming a set of operators with the same generative parameters but time $t$ allowed to vary in $[0,3]$ }
% \vskip -0.2cm
\label{ADRSI}
\end{figure}

\subsection{Burger's equation}\label{BurgerSI}

\begin{figure}[h!]
\centering
\includegraphics[width=0.8\textwidth]{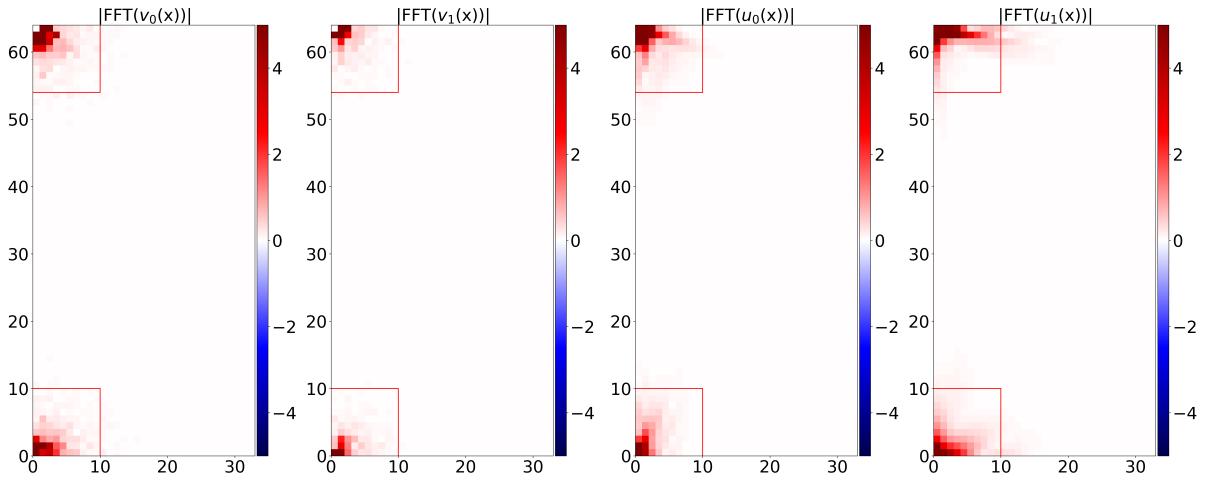}
\caption{Magnitude of the complex coefficients of the Fourier transform of an exemple pair of input and output functions $(\bm{v}(\vec{x}),\bm{u}(\vec{x}))$ in the two coordinates dimension. For every pair, the majority of the signal lies in the two the red quadrants.}
\vskip -0.2cm
\label{ADRSI}
\end{figure} 

\textbf{Generation} $\diamond$ In order to produce the meta-datasets of our second experiment, we use the $\Phi$Flow library  \citep{holllearning} that allows for batched and differentiable simulations of fluid dynamics and available at \url{https://github.com/tum-pbs/PhiFlow}. Following the same methodology as experiment 1, we generate batches of the state evolution of random functions $(\bm{v}_i):\mathbb{R}^2\mapsto\mathbb{R}^2$ defined on the domain $\Omega=[0,1]^2$ at a resolution of $64\times64$ through different parametrization of equation \eqref{Burger}. We form a meta training set of 200 operator datasets for different parameters $\bm{\nu} \in [0.1,0.5]$ each of cardinality $I=100$, and meta testing set of 200 different operator datasets with the same cardinality. Here, we consider vector fields input functions $\bm{v}(\vec{x})$ whose coordinates $(\bm{v}_1(\vec{x}),\bm{v}_2(\vec{x}))$ are drawn each from a two-dimensional zero-mean Gaussian random fields with uniform exponential covariance function and correlation length $l=0.125$. 

\textbf{Discrete Fourier representation} -- Since we are dealing with high-dimensional inputs, we perform kernel regression on the 2D fast Fourier transforms of our model. To reduce further dimensionality, since the FFT of a real signals is Hermitian-symmetric, we pass as input to our model only the flattened $10\times10$ upper and lower quadrants of the Fourier transform coefficients, since we verified that those are sufficient to reconstruct the signal at relative error level of $1e-5$. (We present examples of the 2D FFT of our signal.) After regression, we reconstruct our model estimate in the spatial domain at the desired $64\times64$ resolution and train for the $L_2$ distance against ground truth.

\begin{figure}[h!]
\centering
\includegraphics[width=0.9\textwidth]{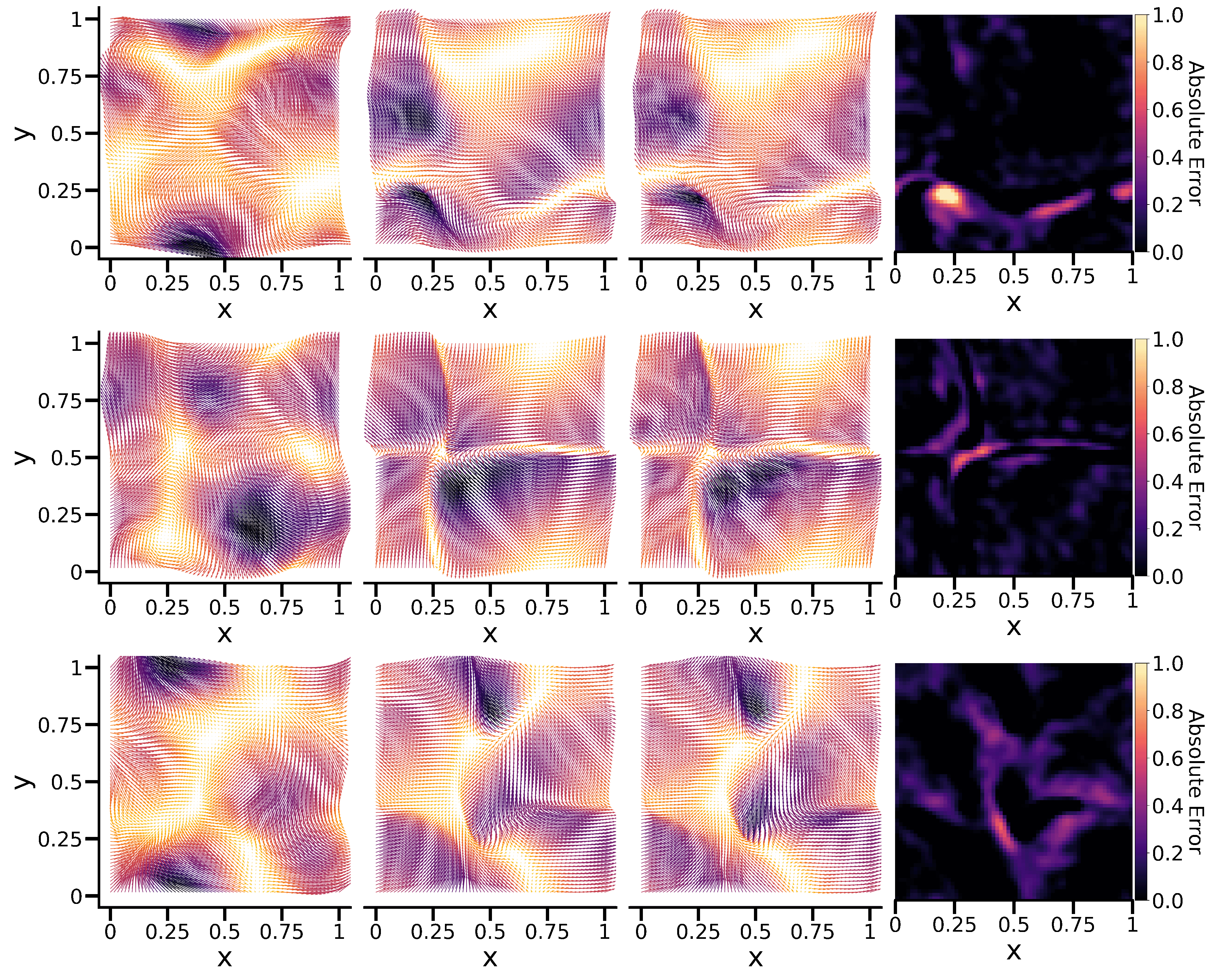}
\caption{Illustrative examples of initial $(t=0)$, target $(t=10)$ and \textit{Transducer} estimation of the vector field $s(\bm{\vec{x}},t)$ discretized at resolution $64\times64$ over the domain $[0,1]^{2}$ for the Burger's equation experiment. The last panel represents absolute error compared to ground truth. }
\vskip -0.2cm
\label{ADRSI}
\end{figure}

\textbf{Spectral clustering} $\diamond$ As a baseline for the outlier detection experiment, we used the spectral clustering algorithm \citep{1238361} implemented in the \texttt{Scikit-learn} on the same FFT preprocessing transformation of the output elements $(\bm{u}_i)_{i\leq I}$ that is discussed above and specifying the number of clusters $C=2$. We tried to tune the clustering algorithm in the embedding space either using K-means or a kernel formulation. The tested variations yielded no significant difference in performance. 

\subsection{Climate modeling}

\textbf{ViT modification} $\diamond$ In order to tackle the high-resolution climate modeling experiment, we take inspiration from \citet{Fourcastnet}, which combines neural operators with the patch splitting method of Vision Transformer (ViT) \citep{dosovitskiy2020vit}. Specifically, we split input and output functions into patches of size $40\times40$. Since both models operations preserves dimensionality, we interleaves \textit{Transducer} layers that apply kernel transformations $\kappa_{\theta}$ along the batch dimension with ViT layers performing spatial attention on the set of patched output function representations $(\bm{u}_i)$. We drop positional encoding but reduce spatial attention to the neighboring patches for each patch position through masking. We compare this bi-attentional model to a vanilla ViT model that learns by induction a single map from temperature $\mathcal{V}$ to pressure $\mathcal{U}$. We double the depth of this baseline to $L=12$, in order to match number of trainable parameters.

\textbf{Data} $\diamond$ We take our data from ERA5 reanalysis \citep{ERA5}, that is freely available on the Copernicus \url{https://cds.climate.copernicus.eu/cdsapp#!/dataset/reanalysis-era5-land?tab=overview}. Surface and temperature pressure are re-gridded from a Gaussian grid to a regular Euclidean grid using the standard interpolation scheme provided by the Copernicus Climate Data Store (CDS) to form 2D fields that we further interpolate in the longitude dimension to obtain images of size $720 \times 720$. Although the ERA5 possess hourly estimates, we subsample the dataset by considering only measurement at 12:00am UTC every day.
% We show in figure \ref{seasonality} a clear seasonal trend in the data that we want to capture through our model.  

\textbf{Training} $\diamond$ As mentioned in the main text, we trained our model to predict variables for 5 days randomly sampled from a 20-day window and condition the \textit{Transducer} with remaining 15 days. We do not explore larger settings due to GPU memory constraints.

\subsection{MNIST-like dataset classification}

\textbf{Training} $\diamond$ We report results from \citet{kirsch2022generalpurpose} for baselines and train and evaluate our model on datasets versions provided by the \texttt{torchvision} library. For this version, we directly treat the images inputs $(\bm{v}_i)_i$ as 784-dimensional vectors and the outputs $(\bm{u}_i)_i$ as 10-dimensional vectors. We do not perform intermediary non-linear transformations $G_{\bm{\theta}}^{\ell}$ for the outputs representations. We haven't performed extensive hyper-parameter search for this experiment in terms of learning rate, head dimensions or kernel expression but simply noted that a deeper 4-layer version of the model was giving similar performance results.

\bibliography{biblio}

% \bibliographystyle{apalike}
% \setcitestyle{authoryear,open={(},close={)}}
% \bibliography{biblio}

\end{document}